\documentclass[10pt,twocolumn,letterpaper]{article}

\usepackage{3dv}
\usepackage{times}
\usepackage{epsfig}
\usepackage{graphicx}
\usepackage{amsmath}
\usepackage{amssymb}

\newcommand{\boldparagraph}[1]{\vspace{0.15cm}\noindent{\bf #1:} }
\newcommand{\centered}[1]{\begin{tabular}{l} #1 \end{tabular}}

\usepackage{longtable}
\usepackage{capt-of}
\usepackage{hhline}
\usepackage{bm}
\usepackage{algorithmic}
\usepackage{algorithm}
\usepackage{siunitx}
\usepackage{multirow}
\newcommand\extrafootertext[1]{%
    \bgroup
    \renewcommand\thefootnote{\fnsymbol{footnote}}%
    \renewcommand\thempfootnote{\fnsymbol{mpfootnote}}%
    \footnotetext[0]{#1}%
    \egroup
}
\usepackage{units}
\usepackage{makecell}

\newcommand\w{0.180}
\newcommand\ww{0.32}

\usepackage{enumitem}

\usepackage[pagebackref=true,breaklinks=true,colorlinks,bookmarks=false]{hyperref}

\threedvfinalcopy 

\def\threedvPaperID{60} 

\ifthreedvfinal\fi
\begin{document}

\title{NeuralBlox: Real-Time Neural Representation Fusion\\for Robust Volumetric Mapping}

\author{Stefan Lionar\thanks{Authors share first authorship.}
\qquad Lukas Schmid\footnotemark[1]
\qquad Cesar Cadena
\qquad Roland Siegwart
\qquad Andrei Cramariuc \\

Autonomous Systems Lab, ETH Zürich, Switzerland \\
{\tt\small splionar@gmail.com \quad \{schmluk, cesarc, rsiegwart, crandrei\}@ethz.ch}
}

\maketitle
\thispagestyle{empty}

\begin{abstract}
We present a novel 3D mapping method leveraging the recent progress in neural implicit representation for 3D reconstruction. Most existing state-of-the-art neural implicit representation methods are limited to object-level reconstructions and can not incrementally perform updates given new data. In this work, we propose a fusion strategy and training pipeline to incrementally build and update neural implicit representations that enable the reconstruction of large scenes from sequential partial observations. By representing an arbitrarily sized scene as a grid of latent codes and performing updates directly in latent space, we show that incrementally built occupancy maps can be obtained in real-time even on a CPU. Compared to traditional approaches such as Truncated Signed Distance Fields (TSDFs), our map representation is significantly more robust in yielding a better scene completeness given noisy inputs. We demonstrate the performance of our approach in thorough experimental validation on real-world datasets with varying degrees of added pose noise. 
\end{abstract}
\vspace{-5mm}

\section{Introduction}
Mapping the 3D volume of an environment during online operation is a fundamental capability required for various robotic tasks, ranging from autonomous navigation~\cite{tsardoulias2016review} to mobile manipulation~\cite{bohren2011towards}. Typically, volumetric maps are built by fusing multiple range measurements into grid-based occupancy or signed distance maps~\cite{hornung2013octomap, oleynikova2017voxblox}. However, such measurements are often affected adversely by uncertainties arising from sensor noise and pose estimation errors~\cite{thrun2002robotic}. To ensure safety during robot operation, \eg obstacle avoidance, these uncertainties must be accurately taken into account in the 3D maps.

\begin{figure}[ht!]
\begin{center}
\begin{tabular}{@{}c@{\hspace{0.2em}}c@{\hspace{0.2em}}c@{}}
\includegraphics[width=0.155\textwidth]{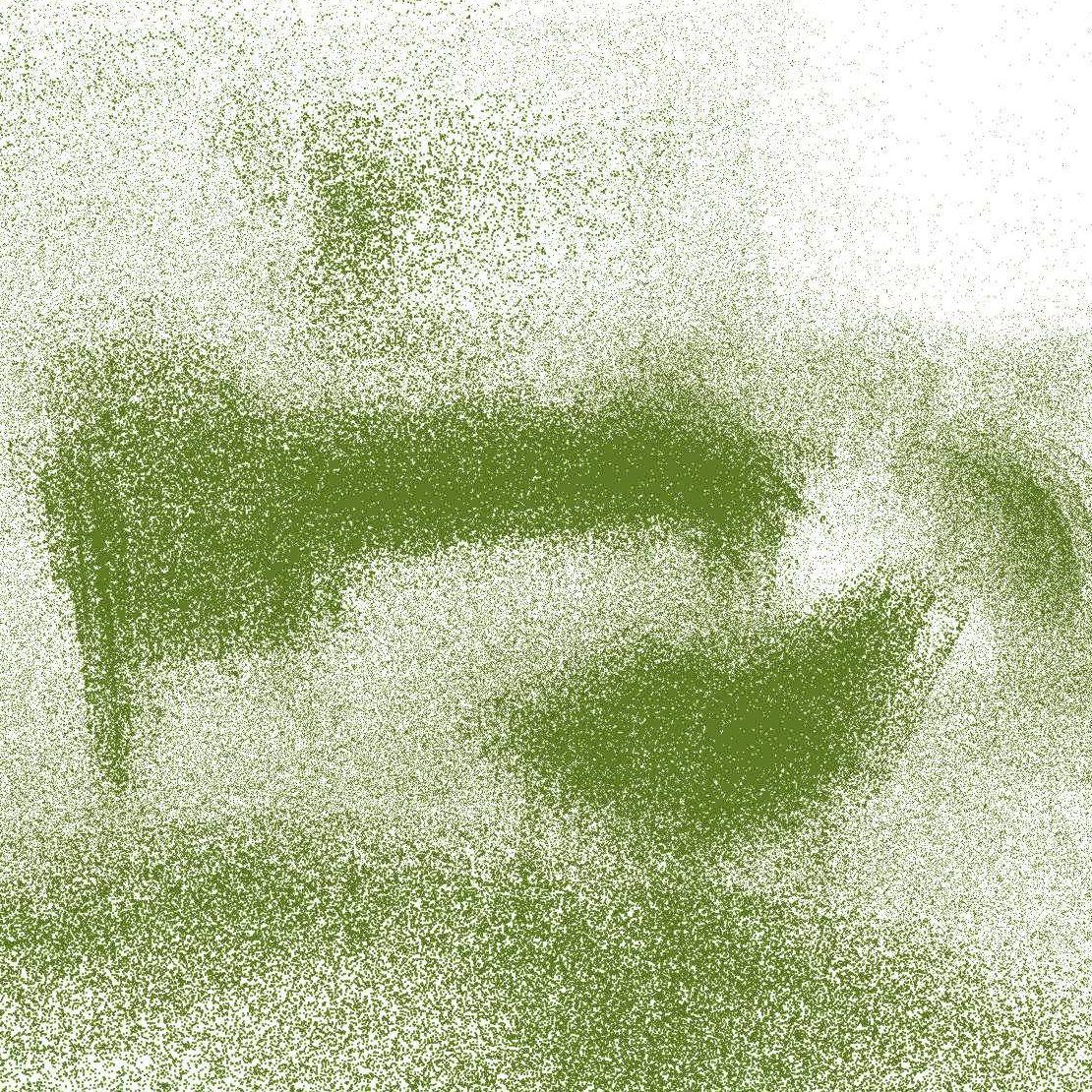} & \includegraphics[width=0.155\textwidth]{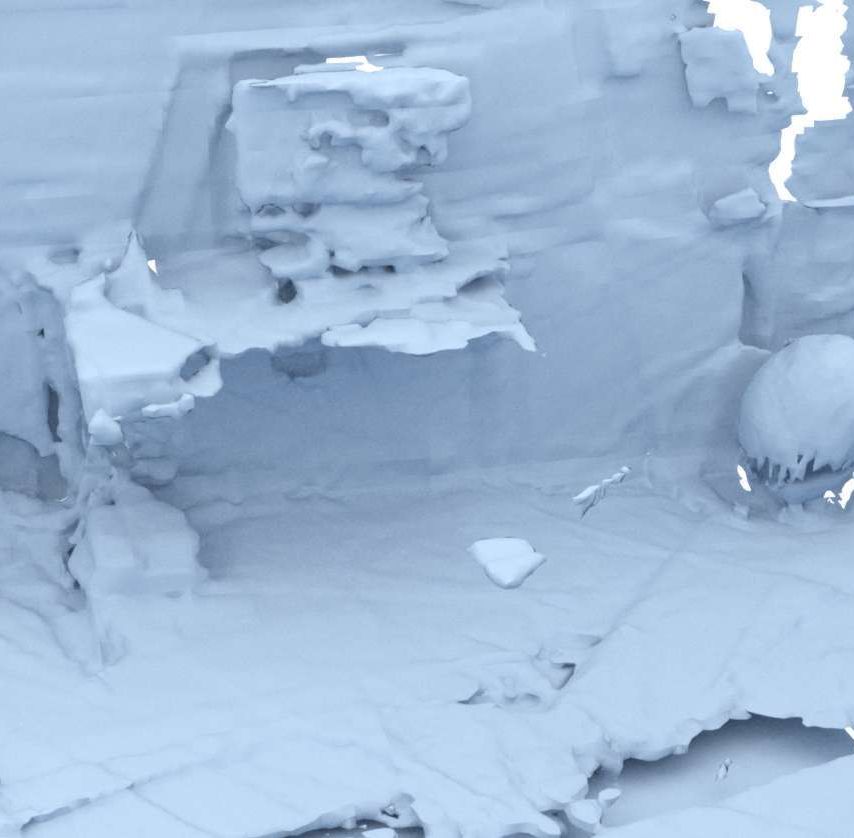} & \includegraphics[width=0.155\textwidth]{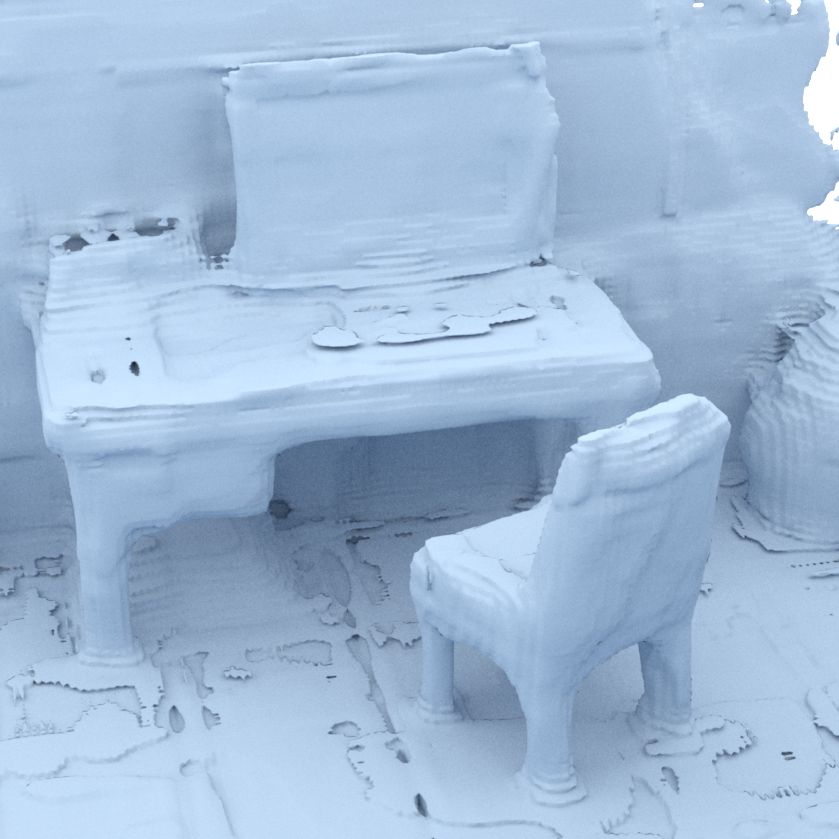} \\
\small{Accumulated inputs} & \small{TSDF~\cite{oleynikova2017voxblox}} & \small{Ours} \\

\end{tabular}
\end{center}
    \caption{Reconstructions of sequential observations with high pose uncertainties. Our method leverages shape priors and geometric context for the fusion, resulting in a robust reconstruction given noisy input.} \vspace{-1em}
    \label{fig:teaser}
\end{figure}

Classical approaches for robotic volumetric mapping often rely on recursive updates of occupancy probabilities~\cite{hornung2013octomap} or Signed Distance Functions (SDFs)~\cite{curless1996volumetric, oleynikova2017voxblox}, which both depend heavily on hand-tuned parameters for specific sensors, scenes, and uncertainties. These approaches apply their update rules without any geometric or semantic context. Consequently, regular occurrence of noise can heavily afflict the resulting SDFs or occupancy probabilities and break the geometry of reconstructed objects in the map. 

Fig.~\ref{fig:teaser} illustrates this undesirable behavior of a TSDF-based method~\cite{oleynikova2017voxblox}. Given sequential observations with high pose uncertainties, the TSDF-based method fails to reconstruct objects in a scene, increasing the risk of collision of the robot and potentially rendering the map insufficient for manipulation tasks. To overcome this limitation, we propose to use a different map representation to fuse range measurements and provide robust volumetric mapping amidst measurement uncertainties. Motivated by the recent progress in neural implicit representation for 3D reconstruction~\cite{park2019deepsdf, chibane2020implicit, peng2020convolutional, lionar2021dynamic, atzmon2020sal, xu2019disn}, we consider representing scenes implicitly as a composition of latent codes that are sequentially encoded from the inputs. These latent codes can then be decoded into other representations such as occupancy grids or SDFs during operation. This enables a broad variety of robotic path planning or interaction algorithms to directly interface with and operate in real-time on our new implicit volumetric mapping framework.

In this work, we propose a system that given sparse unoriented point clouds from \eg a LiDAR or RGBD sensor produces occupancy probabilities as the final output. We represent the map as a dynamically growing grid of large voxels, where each voxel contains a latent code representing the local geometry. Given new measurements, we encode the points into latent implicit representations and perform the map update directly in latent space, instead of updating only the occupancy probabilities. We show that our 3D mapping approach can accurately capture noisy measurements and preserve the overall geometry of the scenes better than classical approaches, as shown in Fig.~\ref{fig:teaser}. In particular, we show that our method achieves high recall, \ie it represents the majority of the scene structure in the map even when subject to noise, which is highly desirable for safe robot operation.

In summary, our contributions are as follows:
\begin{itemize}[noitemsep]
  \item We propose a novel method for online volumetric mapping of arbitrary-sized environments based on neural implicit representations and provide a training strategy to incrementally fuse geometric information from sequential point cloud scans directly in latent space.
  \item We show the flexibility of the presented method to operate on sparse inputs from arbitrary range sensors, generalizing to varying configurations without retraining, and operate in real-time even on just a CPU.
  \item We demonstrate in thorough experimental evaluations that our method can build maps with significantly better scene completeness in the presence of pose estimation errors.
  \item We make the code available as open-source for the benefit of the community\footnote{\url{https://github.com/ethz-asl/neuralblox}}.
\end{itemize}

\section{Related Work}

\boldparagraph{Neural implicit representation} Recently, many works have proposed methods to learn continuous implicit functions for shape representations. This line of methods has shown superior quality and flexibility compared to explicit representations such as voxels~\cite{wu2016learning, wu20153d, hane2017hierarchical}, points~\cite{fan2017point, wangpoint}, and meshes~\cite{gkioxari2019mesh,papier,human}. The pioneering implicit representation methods typically encode the geometric input data into a single latent code, which can then be used to condition the decoder to predict the occupancy or SDF values of query locations in continuous space~\cite{mescheder2018occupancy, park2019deepsdf}. Due to the use of a single latent code, many of these methods are prone to losing low-level details and do not scale well to large scenes. This has been improved in subsequent works by using more sophisticated structures composed of multiple latent codes, such as planes~\cite{peng2020convolutional, lionar2021dynamic}, grids~\cite{chibane2020implicit, peng2020convolutional} and Gaussians~\cite{genova2019learning, genova2020local}. Several recent works also explore broader applications, such as novel view synthesis~\cite{sitzmann2019scene, mildenhall2020nerf, martin2021nerf, pumarola2021d, chibane2021stereo}, representing articulated objects~\cite{deng2020nasa, mihajlovic2021leap}, and learning implicit representation only from 2D observations~\cite{liu2019learning, liu2020dist, niemeyer2020differentiable}.

While most of these works focus on the reconstruction of single objects, Convolutional Occupancy Networks~\cite{peng2020convolutional} and Neural Distance Fields~\cite{chibane2020ndf} demonstrate the ability to reconstruct large scenes. However, their methods work offline requiring complete point clouds of the scene beforehand. Concurrently, Jiang~\etal~\cite{jiang2020local} and Chabra~\etal~\cite{chabra2020deep} proposed methods that represent a scene as coarse latent grid structures which can be decoded to implicit surfaces. However, their methods need to perform optimization for each grid during inference and thus are not suitable for online reconstruction. A common limitation of the aforementioned methods is that they require complete input data and do not consider online updates of the shape representations given sequential partial observations for large-scale mapping.

\boldparagraph{Classical volumetric mapping} The two common representations for volumetric mapping are occupancy and TSDF maps. In OctoMap~\cite{hornung2013octomap}, a 3D scene is segmented into hierarchical octree nodes, where each node stores an occupancy probability to build memory-efficient occupancy maps. While having a straightforward probabilistic update equation, OctoMap requires hand-tuning of the sensor uncertainties to achieve accurate results. 

Similarly, inconvenient hand-tuning of parameters is also required for classical SDF-based approaches that are mostly built upon the TSDF fusion method proposed by Curless and Levoy~\cite{curless1996volumetric}, where uniform grids of SDF values are updated in a weighted averaging fashion. This approach is integrated into many online scene reconstruction frameworks, such as KinectFusion~\cite{izadi2011kinectfusion}, BundleFusion~\cite{dai2017bundlefusion}, Voxgraph~\cite{reijgwart2019voxgraph}, Voxel Hashing~\cite{niessner2013real}, or InfiniTAM~\cite{prisacariu2017infinitam}. However, defining the optimal weighting function for the update rule is a non-trivial task. While Curless and Levoy~\cite{curless1996volumetric} set it to be a constant of 1, KinectFusion~\cite{izadi2011kinectfusion} proposes a weighting function that is proportional to the cosine of angle between the ray from the sensor origin and the normal of the surface. In a sensor model-based approach, Voxblox~\cite{oleynikova2017voxblox} proposes another function using a quadratic weight based on the measured depth. 

All of these approaches perform explicit updates without any shape context and therefore are oftentimes not robust to noisy inputs. In contrast, our method leverages shape priors from training data to more robustly deal with noise or partial observations.

\boldparagraph{Learning-based volumetric mapping} 
More recently, RoutedFusion~\cite{weder2020routedfusion} introduced a learning-based approach for online TSDF updates and weights, resulting in a significant improvement in handling sensor noise. Similar to our approach, NeuralFusion~\cite{weder2021neuralfusion} performs depth fusion directly in a learned latent space and extracts an interpretable representation (TSDF) afterward. However, their fusion approach does not scale well to larger scenes or pose noise.
Inspired by the recent progress in neural rendering~\cite{martin2021nerf}, iMAP~\cite{sucar2021imap} represents a scene using a small multi-layer perceptron that is trained in live operation to predict color and volume density, which can then be converted into an occupancy map. However, iMAP has limited scalability as the entire scene is represented in one single code. Since this code takes time to optimize to match the sensor measurements its usability for \eg online planning is not clear.
In a concurrent work, Huang \etal~\cite{huang2021di} recently presented DI-Fusion, which similarly to our proposed method uses a grid of neural representations that are temporally fused together. As opposed to previous methods that tightly include RGBD camera models or require dense scans for surface normal estimation, our method does not assume any particular distribution of input points and could potentially be extended to other types of range sensors.

\begin{figure*}[!ht]
    \centering
    \includegraphics[width=0.95\textwidth]{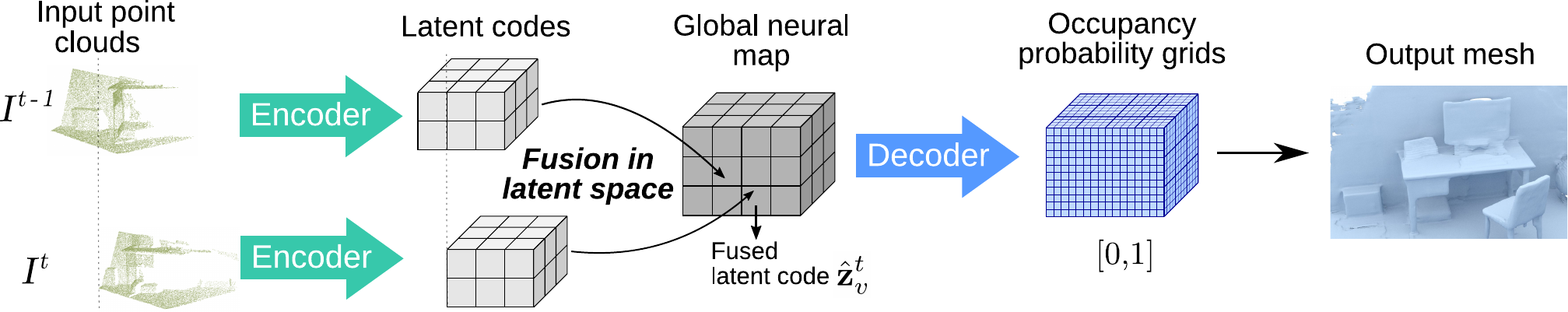}

    \caption{
    Overview of our method. We encode a stream of input points clouds into latent codes and fuse them in latent space to incrementally build a global neural map. This neural map can be decoded at anytime into an occupancy map.
    }
    \label{fig:pipeline}
\end{figure*}

\section{Method}

Fig.~\ref{fig:pipeline} presents an overview of our pipeline. Given a sequence of input point clouds $I^{1:t} \in \mathbb{R}^{3\times N}$, the goal of our method is to produce an occupancy map $o: \mathbb{R}^3 \rightarrow \{\text{occupied}, \text{free}\}$, which contains information about free and occupied space. This map can then be queried at any time step $t$ to enable online robotic tasks such as active path planning or obstacle avoidance. 

Since robots oftentimes operate in an open world setting, we aim to generate an occupancy map of any arbitrarily sized scenes. To accurately model the scene, we partition 3D space into large voxels $\mathbb{V} = \{v\}$, similar to \cite{jiang2020local}. We denote the voxel side length as $d_V$, which is typically within $\SI{0.5}{m}$ to $\SI{1.0}{m}$. In contrast to \cite{jiang2020local}, we dynamically allocate voxels which are inside the view frustum or contain input points to scale to unknown scene sizes. Each voxel contains a latent code $\mathbf{z}_v$, where the index $v$ is used to denote a specific voxel, to represent the geometry within the fixed volume of each voxel. To prevent discontinuities between voxels, we set an overlapping input dimension $d_I (>d_V)$ that defines the input volume of a particular voxel. 

When a new pointcloud $I^t$ arrives, we convert all points that fall into the input volume of each voxel $I^t_v$ into the coordinates of that voxel. Subsequently, we encode $I^t_v$ into a latent code $\mathbf{z}^t_v$ and fuse this shape information into the voxel. This results in fast updates and a compact map representation, where each voxel considers the geometric context within its cell and aggregated shape information from all observations $I_v^{1:t}$.

Afterward, the fused latent codes $\hat{\mathbf{z}}_v^t$ can be decoded via a decoder network into occupancy probabilities for map queries. We set the query dimension $d_q (\geq d_V)$ to construct a query volume, in which occupancy probabilities of points will be queried given the latent code of the corresponding voxel. The larger query volume enables us to smoothen artifacts between neighbouring voxels by interpolating occupancy probabilities in the overlapping regions. Lastly, a high-resolution output mesh can be generated from the occupancy grids using \eg the marching cubes algorithm~\cite{lorensen1987marching}.

We separate the training pipeline into two steps. First, we train the encoder and decoder in a supervised way using the same subset as Choy~\etal~\cite{choy20163d} of ShapeNet~\cite{chang2015shapenet}, a synthetic object-level dataset with the pre-processing steps from Occupancy Networks~\cite{mescheder2018occupancy}. Second, using the trained encoder and decoder, we train a fusion network in a self-supervised manner using a real-world scene dataset~\cite{Park2017} to perform latent code updates.

\subsection{Local Geometry Representation}

Our encoder and decoder mainly follow the architecture of 3D grid Convolutional Occupancy Networks~\cite{peng2020convolutional}, a neural implicit representation approach that is able to capture local spatial context in their implicit representation. Input point clouds are encoded using PointNet~\cite{pointnet} and 3D U-Net~\cite{unet3D}. In the decoding phase, occupancy probabilities of query points in continuous space, $\mathbf{p} \in {\mathbb{R}^3}$, are then predicted using a lightweight decoder network.

\boldparagraph{Architecture adaptation} In the original implementation of Convolutional Occupancy Networks~\cite{peng2020convolutional}, the most compressed representation processed from the input is in the deepest layer of 3D U-Net. However, due to the skip connections in 3D U-Net, this compressed representation cannot be decoded independently from the input. As we are interested in storing the most compressed representation in our map without storing its original input, we modify the 3D U-Net by removing its skip connections. The architecture and training details of our modified encoder and decoder are provided in the supplementary material.

\boldparagraph{Encoder} 
We denote the encoder with weights $\theta_e$ as $f_{\theta_e}$ and input point clouds as $I_v^t$, where no particular structure or density is imposed on $I_v^t$. The latent code $\mathbf{z}_v^t$ is obtained as:

\begin{equation}
    \mathbf{z}_v^t = f_{\theta_e}(I_v^t)
    \label{eq:latent_code}
\end{equation}

\boldparagraph{Decoder} Given the latent code, the goal of the decoder is to predict the occupancy probability of any point $\mathbf{p} \in \mathbb{R}^3$ within the query volume of voxel $v$. The decoder $g_{\theta_d}$, with weights $\theta_d$ processes a latent code $\mathbf{z}_v^t$ and generates the occupancy probability at the query point $\mathbf{p}$:

\begin{equation}
    P_{occ}(\mathbf{p}|\mathbf{z}_v^t) = g_{\theta_d}(\mathbf{p}, \mathbf{z}_v^t) ,\quad  P_{occ} \in [0,1]
    \label{eq:occupancy_prediction}
\end{equation}

\subsection{Latent Code Fusion}
\label{sec:fusion}

The goal of the latent code fusion is to update the current map estimate given a new input, \ie, the updated latent code will contain more complete geometrical information from sequential partial observations. Given the trained encoder $f_{\theta_e}$ and decoder $g_{\theta_d}$, we train the fusion network $h_{\theta_f}$ in a self-supervised manner on a real-world scene dataset. During training, we set input dimension $d_I = \SI{1.2}{m}$, voxel size $d_V = \SI{1.0}{m}$ and query dimension $d_q =d_I$ and do not perform boundary interpolation. With the frozen encoder and decoder, we train the fusion network to predict a fused latent code at any time step $t$, denoted as $\hat{\mathbf{z}}^{t}$. An objective function is designed such that $\hat{\mathbf{z}}^{t}$ has similar properties to the latent code encoded from the accumulated input point clouds up to time step $t$, $\mathbf{z}^{\ast t}$. The latent code of voxel $v$ based on the accumulated input point clouds that have fallen into its input volume until time step $t$ is defined as:

\begin{equation}
    \mathbf{z}^{\ast t}_{v} = f_{\theta_e}(I^{1:t}_v)
\label{eq:gt}
\end{equation}

Since we only consider measurements that affect a voxel $v$, we denote the set of relevant time steps as:

\begin{equation}
\mathbb{T}^t_v = \{ \tau \bigm| |I_v^\tau| > 0 \}_{\tau=0,\dots, t}
\end{equation}

To generate the fused latent code prediction, we keep track of previous latent codes in an averaging fashion. This enables a constant memory footprint for each voxel, which stores the summation of latent codes from input point clouds that fall into it:

\begin{equation}
 \bar{\mathbf{z}}_v^t = \sum_{\tau \in \mathbb{T}^t_v} f_{\theta_e}(I^\tau_v)
\end{equation}
and the number of measurements $N_v= |\mathbb{T}^t_v|$.
The fused latent code of voxel is generated by first dividing the summation by $N_v$, then feeding it into the fusion network $h_{\theta_f}$:

\begin{equation}
    \hat{\mathbf{z}}^{t}_{v} = h_{\theta_f}(\frac{\bar{\mathbf{z}}_v^t}{N_v})
\label{eq:pred}
\end{equation}

\boldparagraph{Feature alignment loss} We employ a feature alignment loss to minimize the distance between $\hat{\mathbf{z}}$ and $\mathbf{z}^{\ast}$ in latent space:

\begin{equation}
  \mathcal{L}_{fea} = \frac{1}{|\mathbb{V}|}\sum_{v \in \mathbb{V}}\big\lVert\mathbf{z}^{\ast t}_{v} - \hat{\mathbf{z}}^{t}_{v}\big\rVert_1
  \label{eq:loss feature}
\end{equation}

This feature alignment loss allows the fusion network to correct for errors introduced by the averaging operation if the latent space is not perfectly linear and encourages the fused codes $\hat{\mathbf{z}}_v$ to stay in the domain of the pre-trained encoder and decoder.

\boldparagraph{Reconstruction loss} Additionally, we add a reconstruction loss that minimizes the logit differences of query points $\mathbf{p}$ decoded using $\hat{\mathbf{z}}$ and $\mathbf{z}^{\ast}$, where $\mathbb{P}$ are randomly distributed sample points:

\begin{equation}
  \mathcal{L}_{rec} =
  \frac{1}{|\mathbb{V}| |\mathbb{P}|}\sum_{\mathbf{p}\in\mathbb{P}}\sum_{v \in \mathbb{V}}\big\lVert g_{\theta_d}(\mathbf{p}, \mathbf{z}^{\ast t}_{v}) - g_{\theta_d}(\mathbf{p}, \hat{\mathbf{z}}^{t}_{v})\big\rVert_1
  \label{eq:loss reconstruction}
\end{equation}

\boldparagraph{Backward pass} During training, we always take 8 sequences of input point clouds and then perform the backward pass $(t=8)$. Therefore, the number of updates of a voxel ($N_v$) can range between 1 and 8. However, we found that this training procedure generalizes well to longer sequences. The total loss function is obtained as the summation of the feature alignment and reconstruction loss:

\begin{equation}
  \mathcal{L} =  \mathcal{L}_{fea} + \mathcal{L}_{rec}
  \label{eq:loss total}
\end{equation}

\boldparagraph{Fusion network} The fusion network $h_{\theta_f}$ is composed of two 3D convolution layers with ReLU activations. The architecture details are in supplementary materials. Due to its shallow structure, the fusion process does not add any noticeable runtime to our pipeline.

\subsection{Occupancy Map Generation}
A threshold $\tau_{occ}$ is set to define whether queried points are considered free or occupied space, where points with occupancy probability less than $\tau_{occ}$ are considered free. We uniformly sample query points in a structure of dense 3D grid for each voxel. If a voxel is not yet allocated, it is considered unknown. Otherwise, if a voxel is allocated but has not yet received any point cloud ($N_v=0$), it directly outputs free space when queried. In our scene reconstruction experiments discussed in Sec.~\ref{sec:experiments}, we store the summation of latent codes and $N_v$ for every voxel and only generate the occupancy after the last input scan. In practice, the occupancy can be decoded at any time step by dividing the summed latent codes by $N_v$, predict the fused latent code, and then feed query points as well as the predicted latent code into our decoder. 

\boldparagraph{Boundary interpolation} We set $d_q = d_V + (d_I - d_V) \times 0.5$, giving overlapping regions between adjacent voxels. When a voxel is decoded, it looks up to its neighboring voxels and performs linear interpolations of probabilities on the overlapping regions. A comparison with and without the boundary interpolation is in the supplementary material.

\section{Experiments}
\label{sec:experiments}

We provide the experimental details comparing the performance of our model and a standard TSDF method implemented in Voxblox~\cite{oleynikova2017voxblox}. We mainly evaluate the robustness of every model subject to inputs with varying levels of pose noise. In Sec.~\ref{sec:redwood}, we also provide thorough details of the runtime of our models in different configurations. 

\subsection{Model Comparison}

Without any retraining, we set up two configurations with different voxel size $d_V$ and input dimension $d_I$ during occupancy map generation, as follows:
\begin{itemize}[noitemsep]
    \item Ours (0.5 m): $d_V = \SI{0.5}{m}$, $d_I =  \SI{0.7}{m}$
    \item Ours (1 m): $d_V =  \SI{1}{m}$, $d_I =  \SI{1.2}{m}$
\end{itemize}

A smaller voxel size results in a finer reconstruction at the expense of storing more latent codes and thus slower inference speed. In both configurations, we employ query points structured in a grid with a resolution of $100^3$ points/m$^3$. By including the query points in overlapping regions, this results in $60^3$ and $110^3$ points/voxel for the 0.5 m and 1 m configurations, respectively.

We set the initial 3D grid resolution to 24 voxels, hidden layer size to 32, and 3D U-Net depth to 2. Thus, for each voxel, input point clouds are processed into a latent code $\mathbf{z}_v$ with a dimension of $6 \times 6 \times 6 \times 128$. By corresponding to the same number of bytes used per m$^3$ between our method and TSDF, we benchmark our model with $d_V = \SI{0.5}{m}$ to a TSDF with a voxel resolution of $\SI{0.02}{m}$ and ours with $d_V=\SI{1}{m}$ to a TSDF with a voxel resolution of $\SI{0.04}{m}$. 
For fast runtime, our method can operate accurately using sparse input points, which is further discussed in Sec.~\ref{sec:redwood}.
Thus in our experiments, we only use 10\% of the points per scan as the input to our models. With this input percentage, we empirically set the threshold $\tau_{occ}=0.05$. Additionally, we do not update a latent code if its corresponding input volume receives less than 1\% of the input points. Conversely, we feed 100\% of the point cloud for TSDF.

\subsection{Metrics}

We use the following metrics for our evaluation:
\vspace{-0.5em}

\begin{itemize}[noitemsep]
    \item \textbf{Accuracy} (lower better): mean absolute distance (MAD) of points sampled densely from the predicted mesh to the closest surface on the ground truth mesh. 
    \item \textbf{Completeness} (lower better): MAD of points sampled densely from the ground truth mesh to the closest surface on the predicted mesh. 
    \item \textbf{Recall} (higher better): the percentage of points sampled from the ground truth mesh that have closest absolute distances to the predicted mesh within a threshold $\tau_r$. We use $\tau_r = \SI{0.05}{m}$.   
\end{itemize}

\begin{figure*}[t!]
\begin{center}
\begin{tabular}{ccccc}
\includegraphics[width=0.15\textwidth]{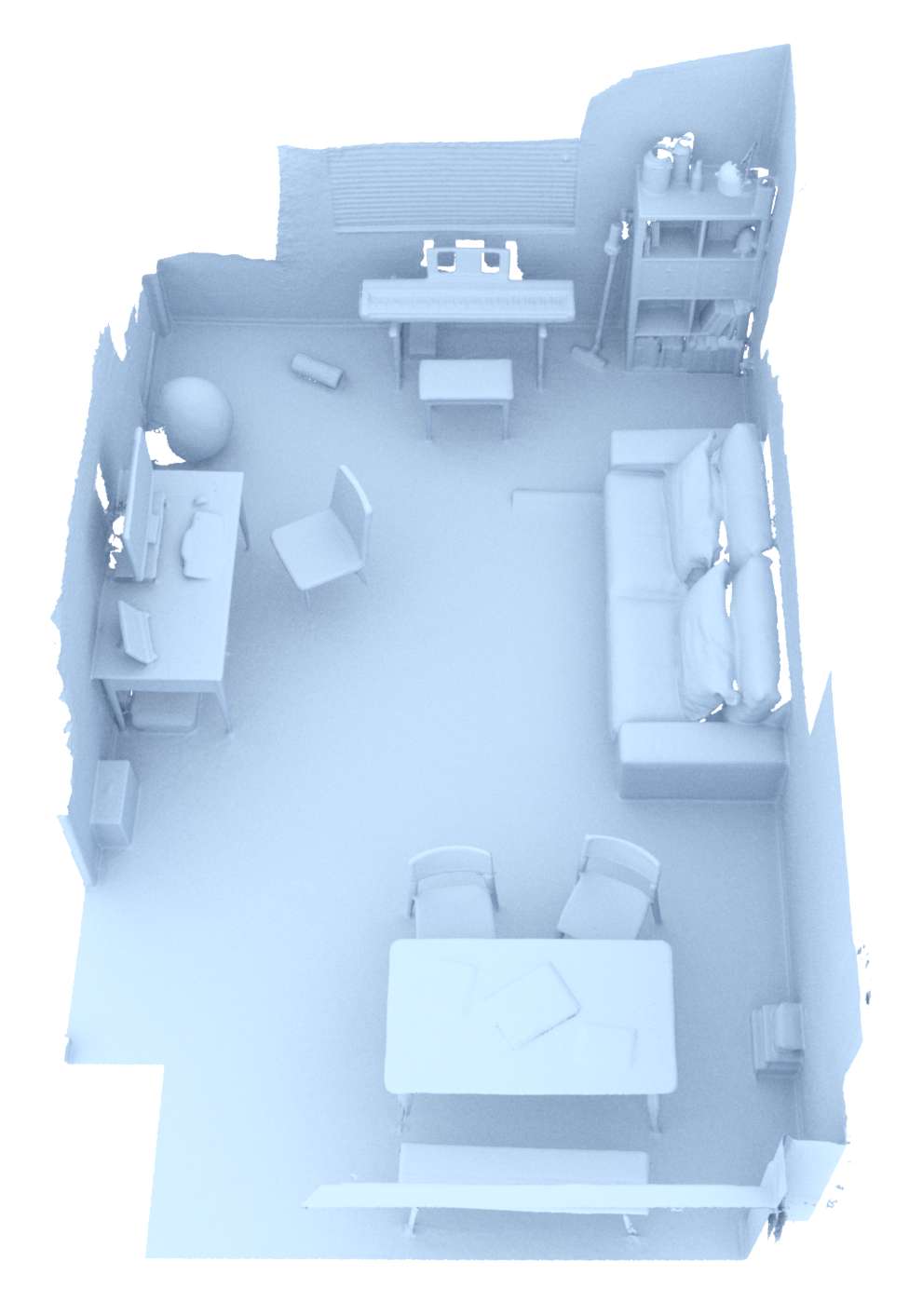} &
\includegraphics[width=0.17\textwidth]{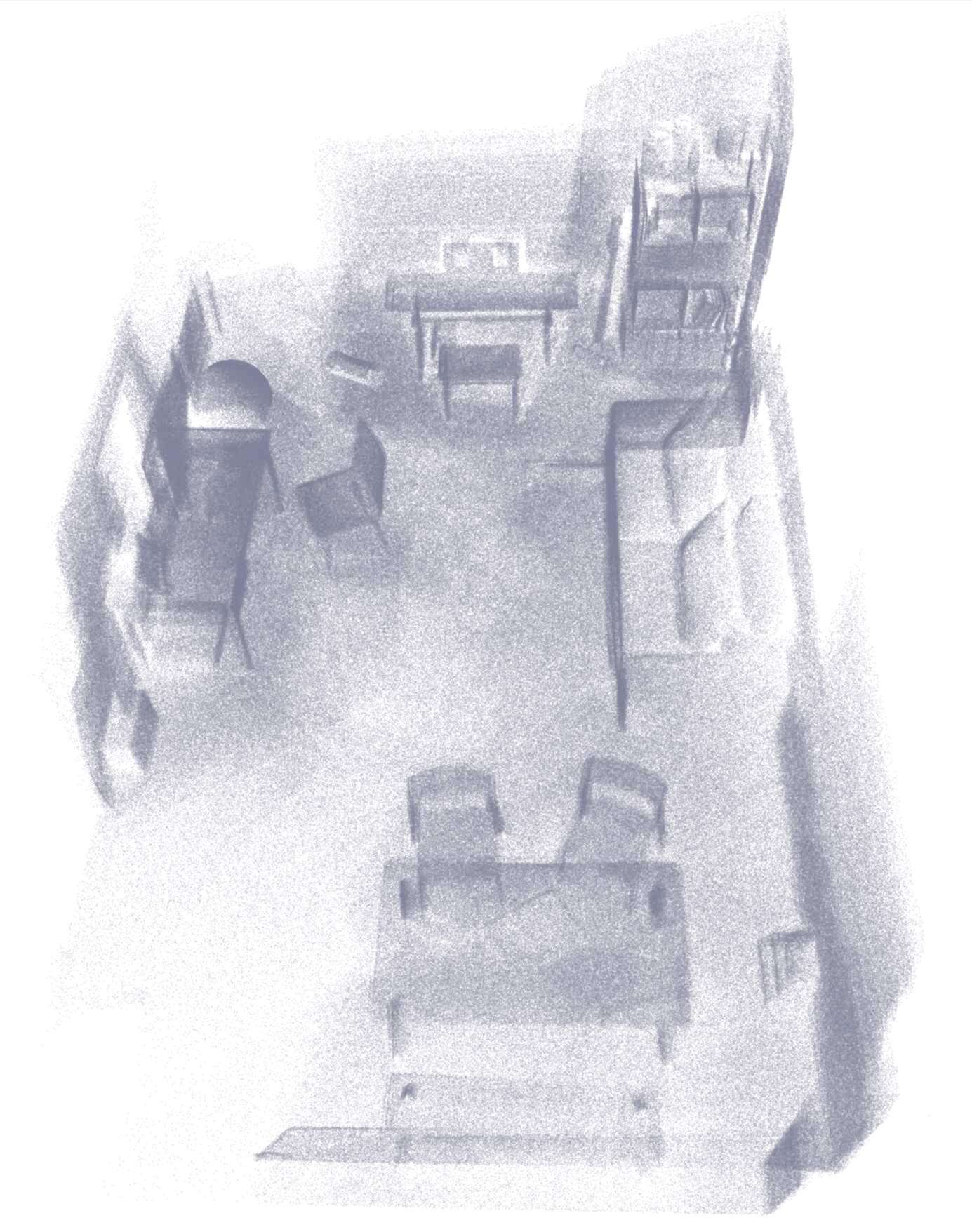} &
\includegraphics[width=0.17\textwidth]{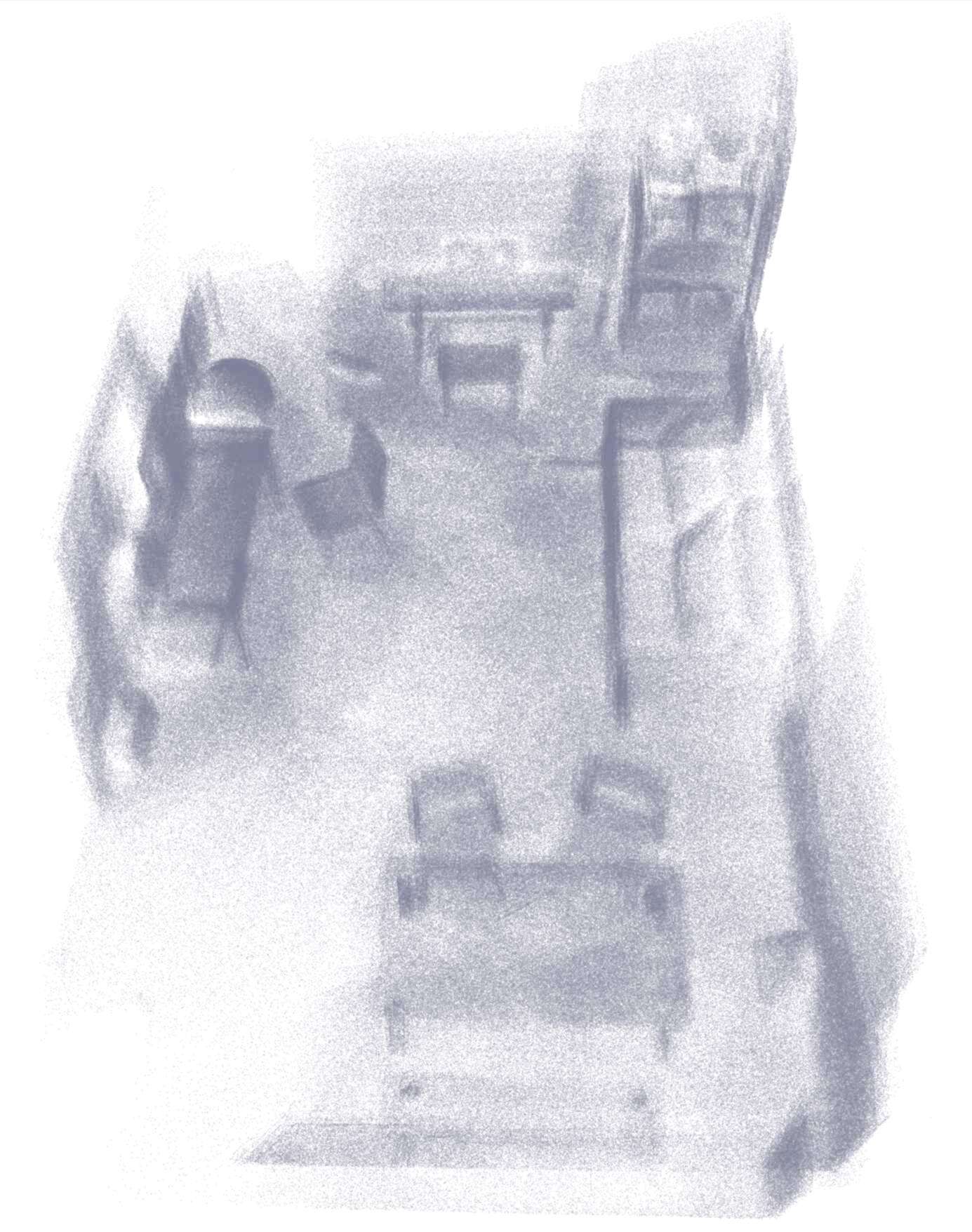} & \includegraphics[width=0.17\textwidth]{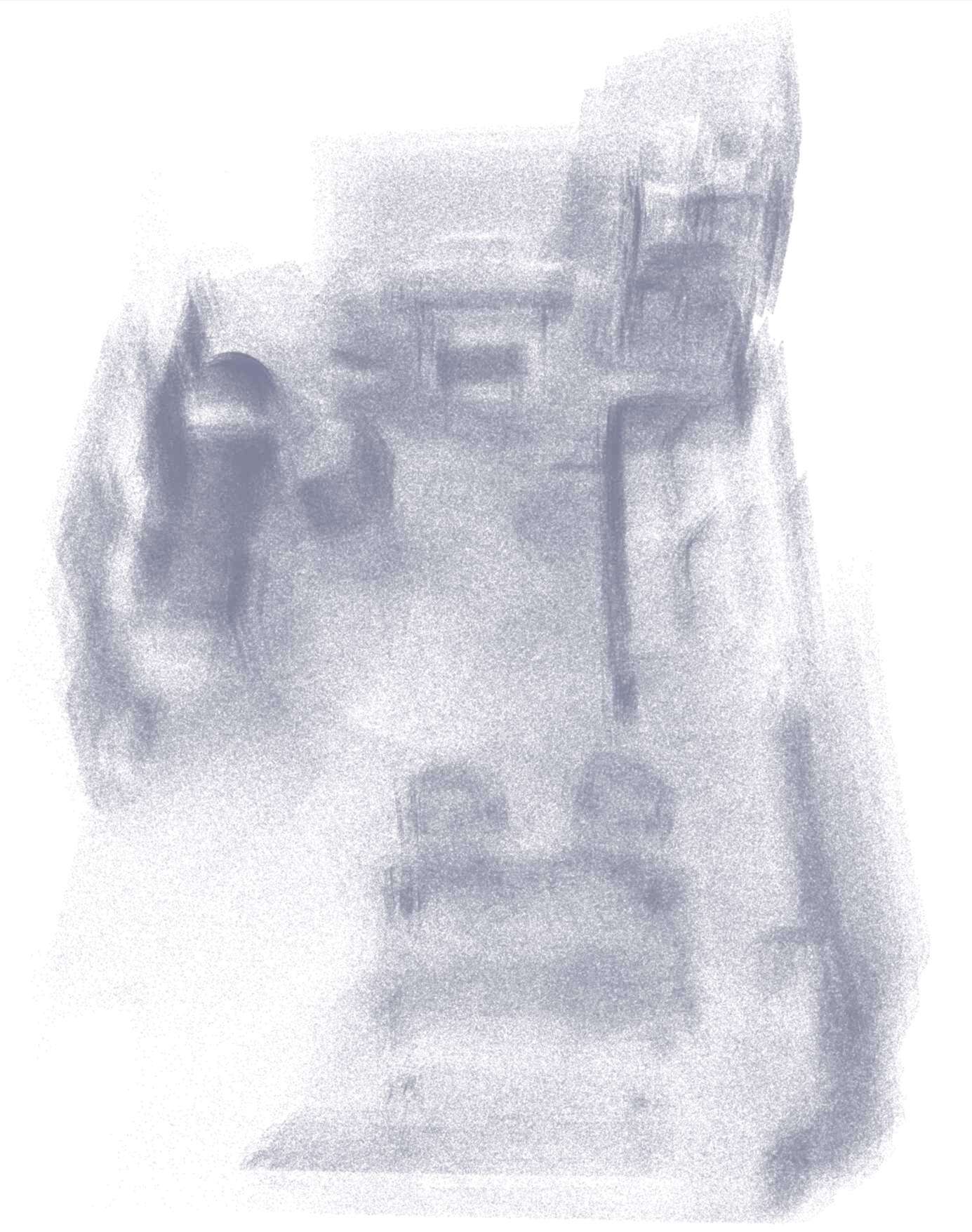} & \includegraphics[width=0.17\textwidth]{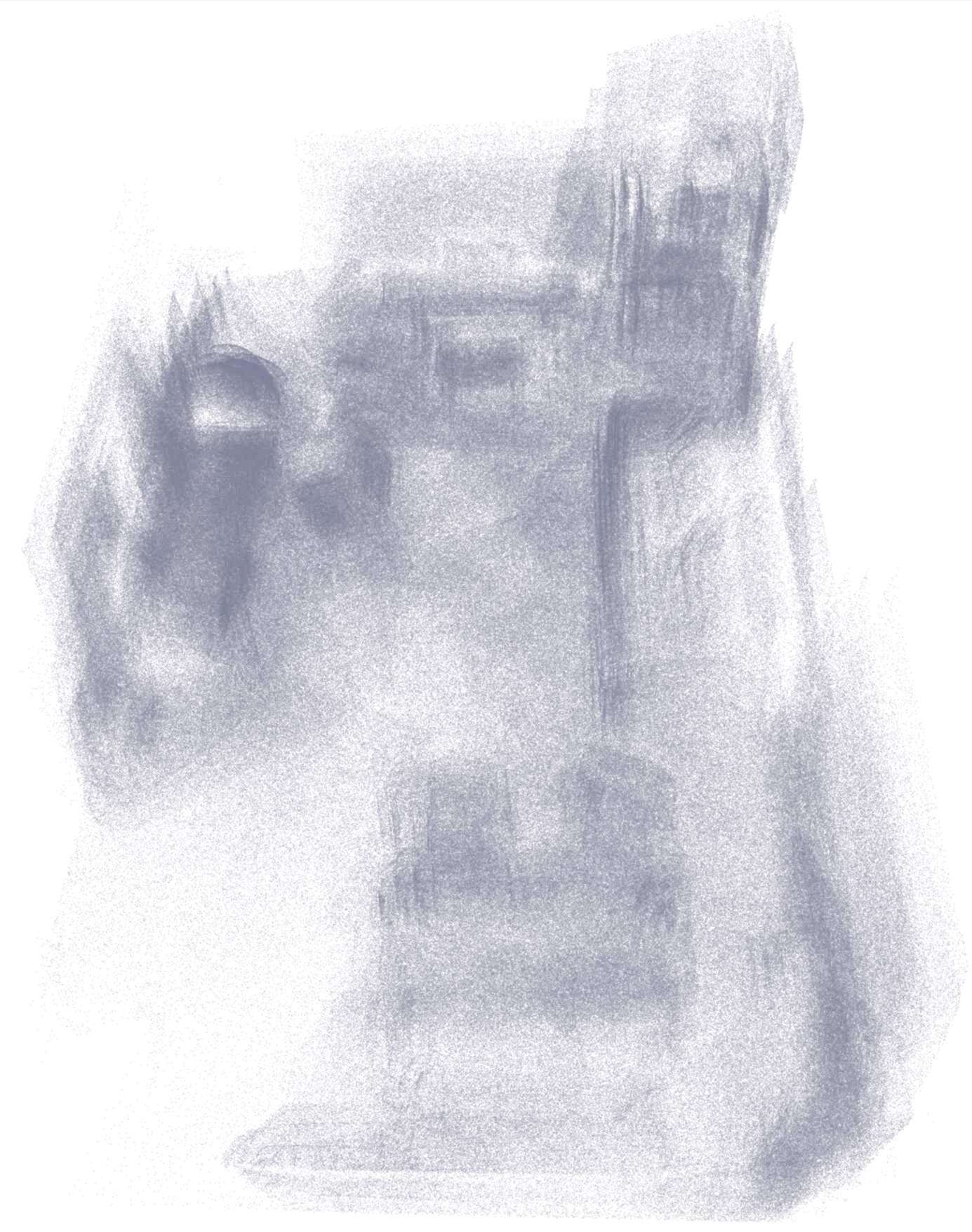} \\
\small{GT~\cite{Park2017}} & \small{$\sigma_T = 0$}  & \small{$\sigma_T = 0.025$} & \small{$\sigma_T = 0.050$} & \small{$\sigma_T = 0.075$} \\  
  
\end{tabular}\\
\end{center}
\caption{Ground truth and sampled accumulated inputs. Different levels of translation noise $\sigma_T$ (m) are applied on $x$ and $y$ axes.} \vspace{-1em}
\label{fig:noise-level}
\end{figure*}

\subsection{Redwood Indoor RGBD Dataset}
\label{sec:redwood}

\boldparagraph{Training} We leverage the Redwood dataset~\cite{Park2017} to train the fusion network. It contains $640\times480$ resolution RGBD sequences of five real-world indoor scenes captured with an Asus Xtion Live camera. We use four scenes, namely \emph{bedroom, boardroom, lobby,} and \emph{loft} as our training data. Nonetheless, we find that using only one scene (\emph{bedroom}) already produces a model that generalizes well. During training, we take 8 depth sequences with alternating separations of 30 and 90 frames in each iteration. Then, we convert them into point clouds with a truncated depth of $\SI{3}{m}$. We sample 25,000 points each scan, which corresponds to about 10\% of total points. Further training details are in the supplementary material.

\boldparagraph{Evaluation} We use the \emph{apartment} scene for evaluation, which consists of approximately 31,910 RGBD frames. We utilize the high-quality mesh, reconstructed offline using the method presented in~\cite{Park2017}, as our ground truth. Evaluations are carried out on the scene thoroughly covered by the first 13,000 frames where we found the ground truth to be most reliable. We take every 30$^{\text{th}}$ frame of the RGBD scans and convert them to point clouds with a truncation distance of 3.0 m as input to the evaluated models. The ground truth mesh is slightly trimmed so that it covers roughly the same area as the input point clouds, as shown in Fig.~\ref{fig:noise-level}. 

To evaluate the robustness of our system given state estimation uncertainties, we add artificial pose errors to the provided extrinsic camera parameters. Since most robots are usually equipped with an Inertial Measurement Unit (IMU), the direction of gravity can typically be well estimated. We thus add noise only to the $x$ and $y$ directions (assuming $z$ is the direction of gravity) of the camera pose. Given the extrinsic camera parameters of a frame:
\vspace{-0.3em}
\begin{align*}
C_{ex} &=  
\left(\begin{array}{cc}
        R & T \\
        0 & 1
    \end{array}\right)
\end{align*}
%
where $R \in SE(3)$ is the rotation matrix and $T \in \mathbb{R}^3$ is the translation vector, we sample two values from a normal distribution with zero mean and standard deviation $\sigma_T$ and add them to the translation components $T_x$ and $T_y$. 

To better illustrate the influence of different levels of $\sigma_T$, we show the sampled accumulated inputs in Fig.~\ref{fig:noise-level}. Quantitative results with varying values of $\sigma_T$ are presented in Tab.~\ref{tab:results-shift} and qualitative visualizations of the reconstructions are presented in Fig.~\ref{fig:qualitative-shift}. In the supplementary material, we also present quantitative and qualitative results of the cases where translation noise is applied to the $x, y,$ and $z$ axes as well as when orientation errors around $z$ (yaw) are present.
 
\begin{table}[h!]
\begin{center}
\resizebox{0.47\textwidth}{!}{%
\begin{tabular}{|c||c|c||c|c||c|c||c|c|}
    \hline
     $\sigma_T$ &\multicolumn{8}{c|}{\emph{\textbf{Left:} TSDF} (0.02 m)~\cite{oleynikova2017voxblox}. \emph{\textbf{Right:}} Ours (0.5 m).} \\
    \cline{2-9}
     (m) & \multicolumn{2}{c||}{\emph{Accuracy} (m)} & \multicolumn{2}{c||}{\emph{Completeness} (m)} & \multicolumn{2}{c||}{\emph{Recall}} & \multicolumn{2}{c|}{\emph{Recall*}}\\
     \hline
      0 & \textbf{0.0175} & 0.0225 & 0.0137 & \textbf{0.0135} & 0.988 & \textbf{0.990} & 0.984 & \textbf{0.986} \\
      0.025 & \textbf{0.0213} & 0.0264 & 0.0202 & \textbf{0.0159} & 0.926 & \textbf{0.989} & 0.888 & \textbf{0.985} \\
      0.050 & 0.0377 & \textbf{0.0372} & 0.0352 & \textbf{0.0216} & 0.756 & \textbf{0.932} & 0.625 & \textbf{0.898} \\
      0.075 & 0.0535 & \textbf{0.0503} & 0.0486 & \textbf{0.0249} & 0.635 & \textbf{0.876} & 0.438 & \textbf{0.809} \\ 
      \hline \hline
      $\sigma_T$  &\multicolumn{8}{c|}{\emph{\textbf{Left:} TSDF} (0.04 m)~\cite{oleynikova2017voxblox}. \emph{\textbf{Right:}} Ours (1 m).} \\
      \cline{2-9}
      (m) & \multicolumn{2}{c||}{\emph{Accuracy} (m)} & \multicolumn{2}{c||}{\emph{Completeness} (m)} & \multicolumn{2}{c||}{\emph{Recall}} & \multicolumn{2}{c|}{\emph{Recall*}}\\
     \hline
      0 & \textbf{0.0206} & 0.0292 & \textbf{0.0160} & 0.0187 & 0.964 & \textbf{0.970} & 0.950 & \textbf{0.957} \\
      0.025 & \textbf{0.0218} & 0.0309 & \textbf{0.0191} & 0.0201 & 0.936 & \textbf{0.964} & 0.906 & \textbf{0.947} \\
      0.050 & \textbf{0.0290} & 0.0367 & 0.0299 & \textbf{0.0245} & 0.817 & \textbf{0.923} & 0.722 & \textbf{0.885} \\
      0.075 & \textbf{0.0446} & 0.0456 & 0.0404 & \textbf{0.0286} & 0.705 & \textbf{0.849} & 0.545 & \textbf{0.768} \\ 
      \hline

    \end{tabular}}\\
    \footnotesize{*The floor is excluded.} \vspace{-0.8em}
    \end{center}
    \caption{Performance of our method on our evaluation set subject to translation noise $\sigma_T$ on the $x$ and $y$ axes. Notably, our method shows high robustness to state estimation noise and is always able to faithfully capture the underlying geometry, as shown by the high recall. This is particularly pronounced for objects of interest, \eg when the floor is excluded. }
    \label{tab:results-shift}
\end{table}

\vspace{-0.3em}

\begin{figure*}[!t]
\begin{center}
\begin{tabular}{@{}c|c@{\hspace{0.2em}}c|c@{\hspace{0.2em}}c@{}}
   & \multicolumn{2}{c|}{\textbf{Small voxels}} & \multicolumn{2}{c}{\textbf{Large voxels}} \\
  $\sigma_T$ (m) & TSDF (0.02 m) & Ours (0.5 m) & TSDF (0.04 m) & Ours (1 m) \\
  \centered{0} & \centered{\includegraphics[width=\w\textwidth]{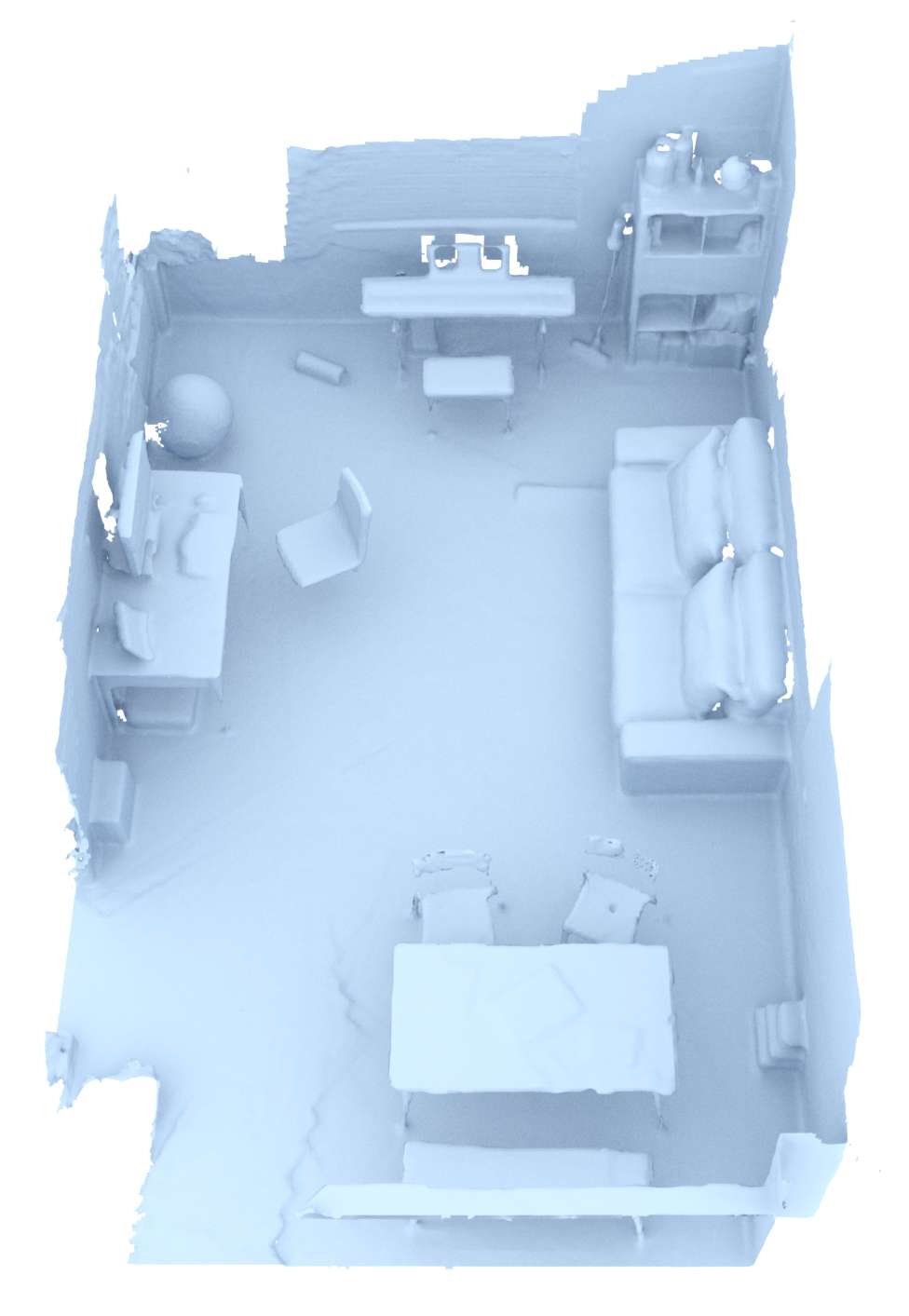}} &  \centered{\includegraphics[width=\w\textwidth]{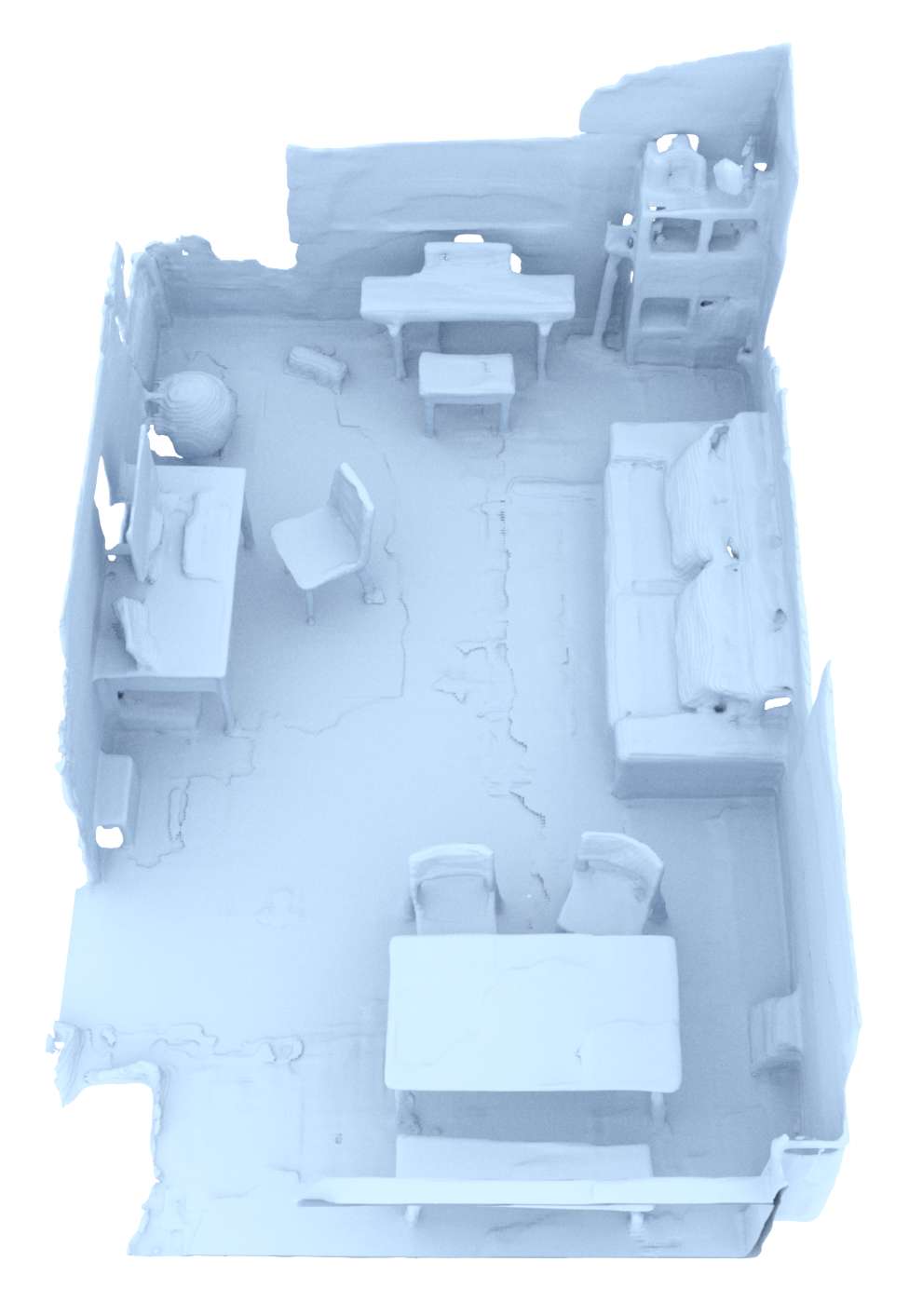}} &  \centered{\includegraphics[width=\w\textwidth]{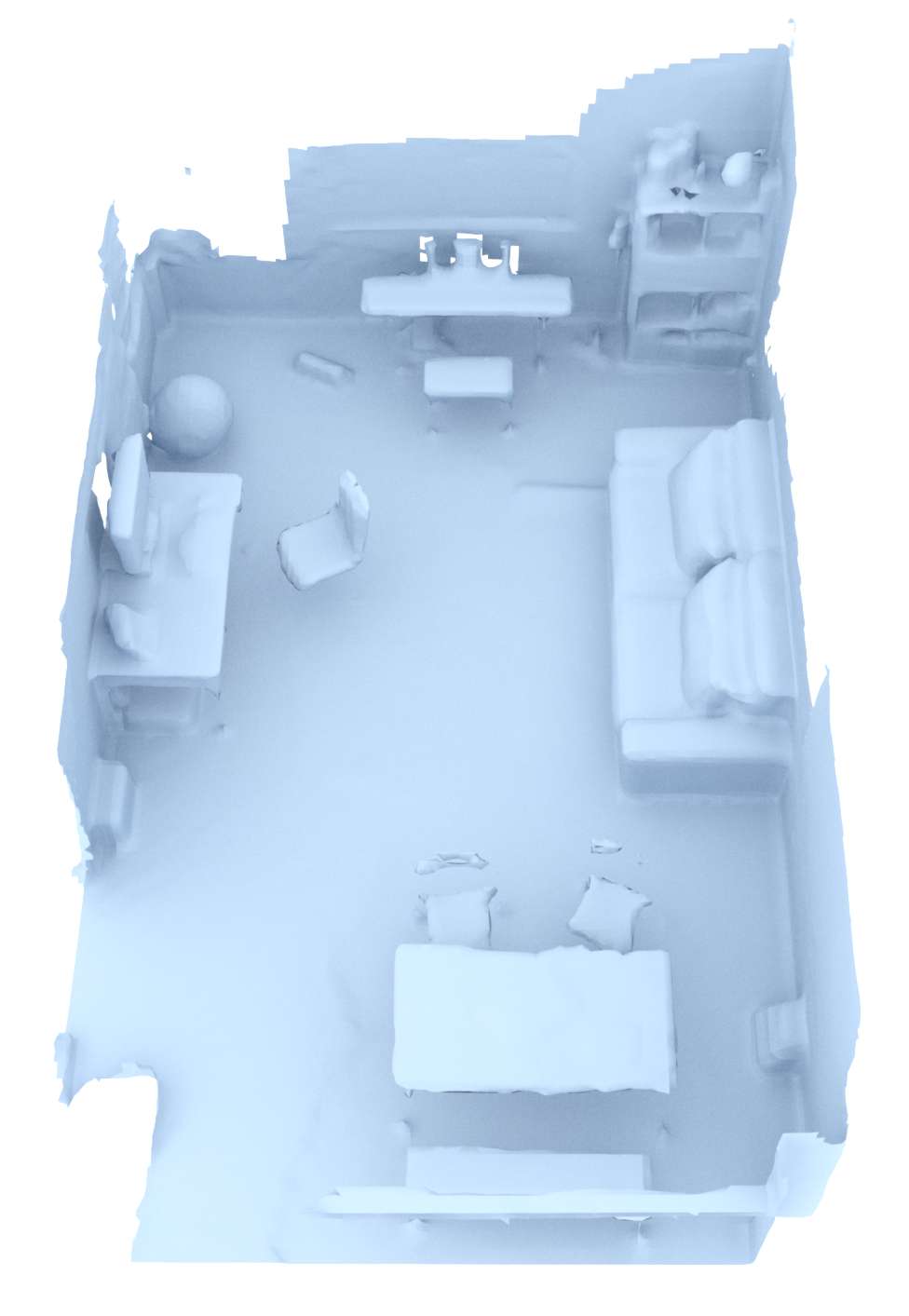}} &  \centered{\includegraphics[width=\w\textwidth]{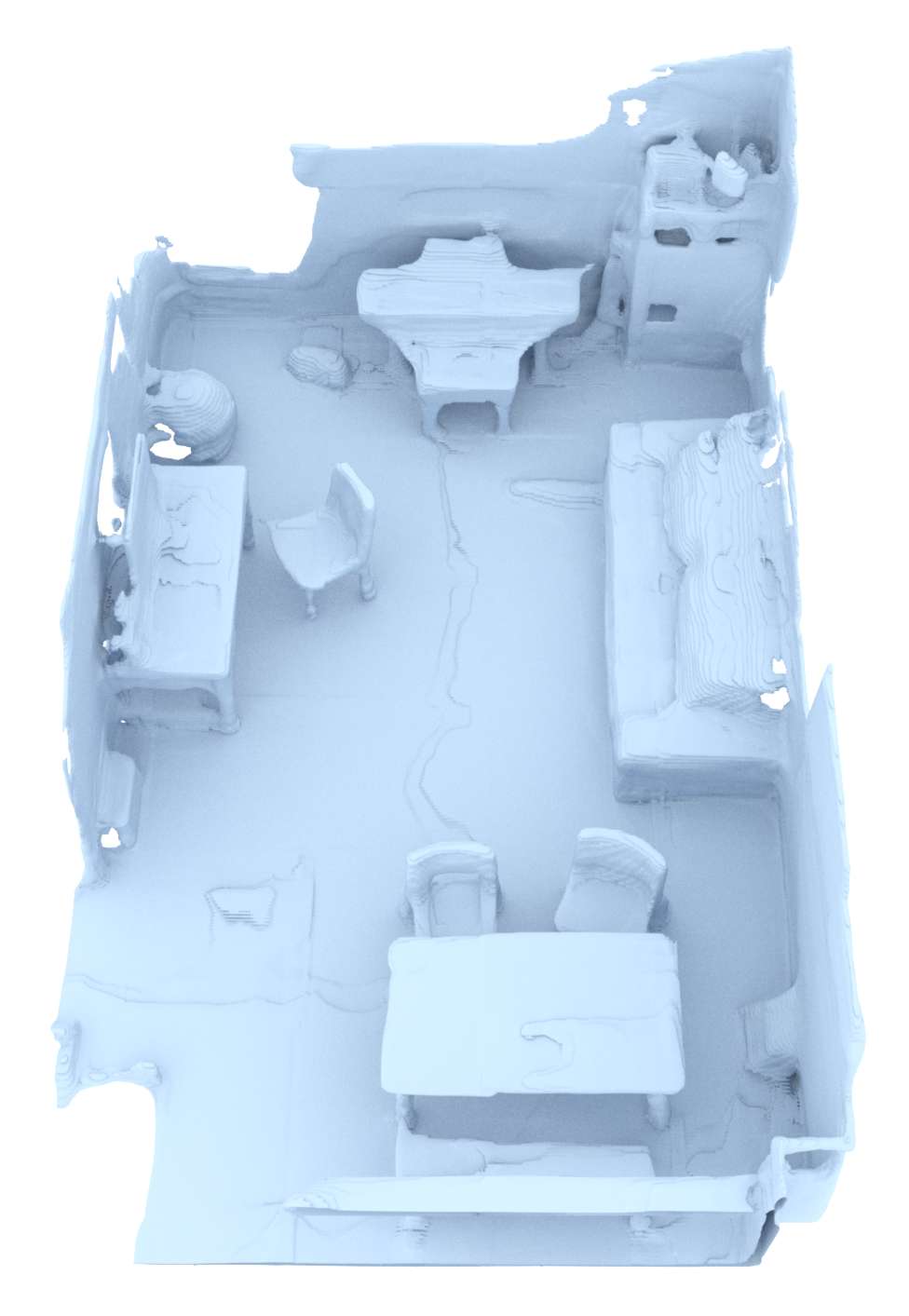}} \\ 
  \centered{0.075} & \centered{\includegraphics[width=\w\textwidth]{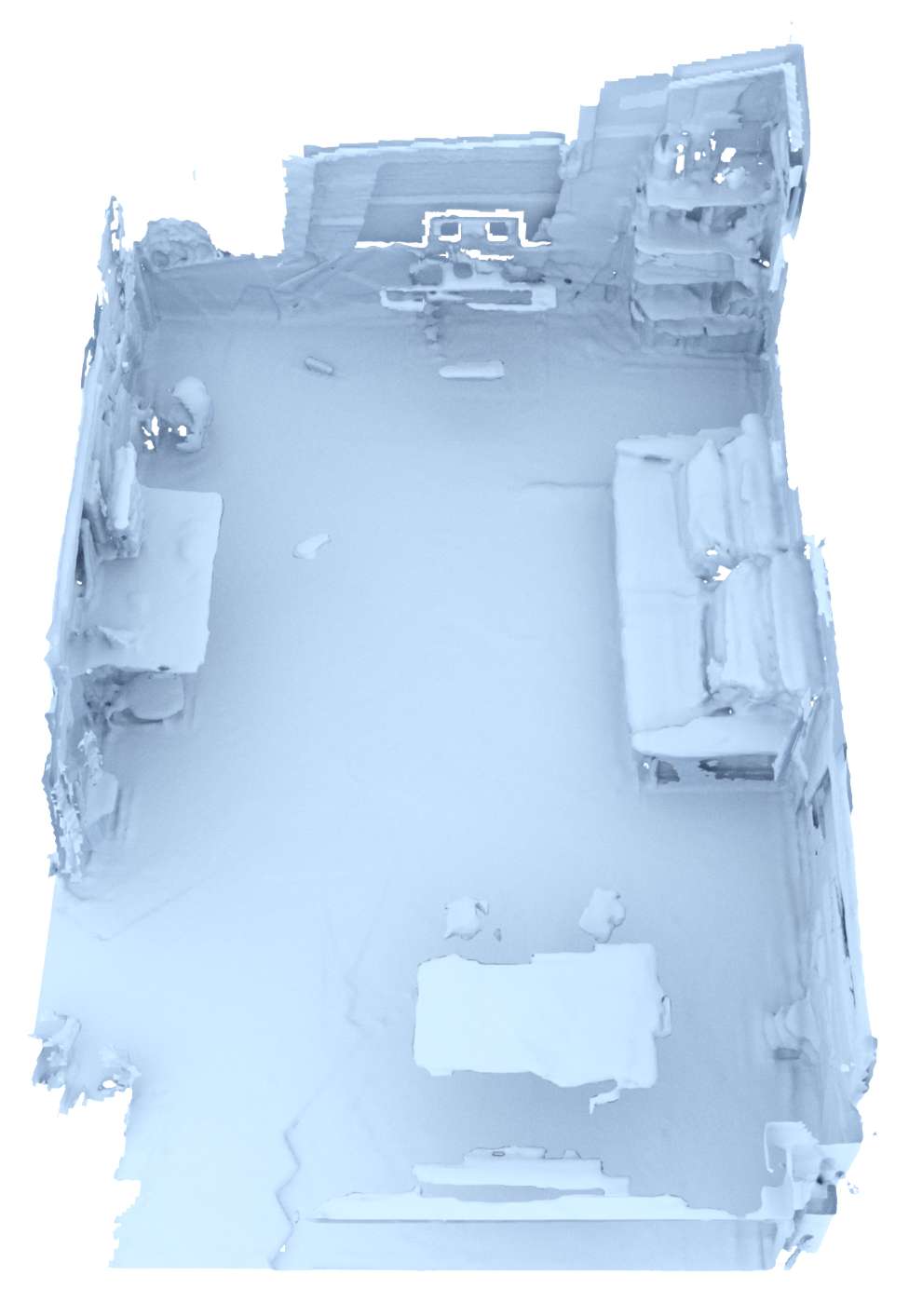}} &  \centered{\includegraphics[width=\w\textwidth]{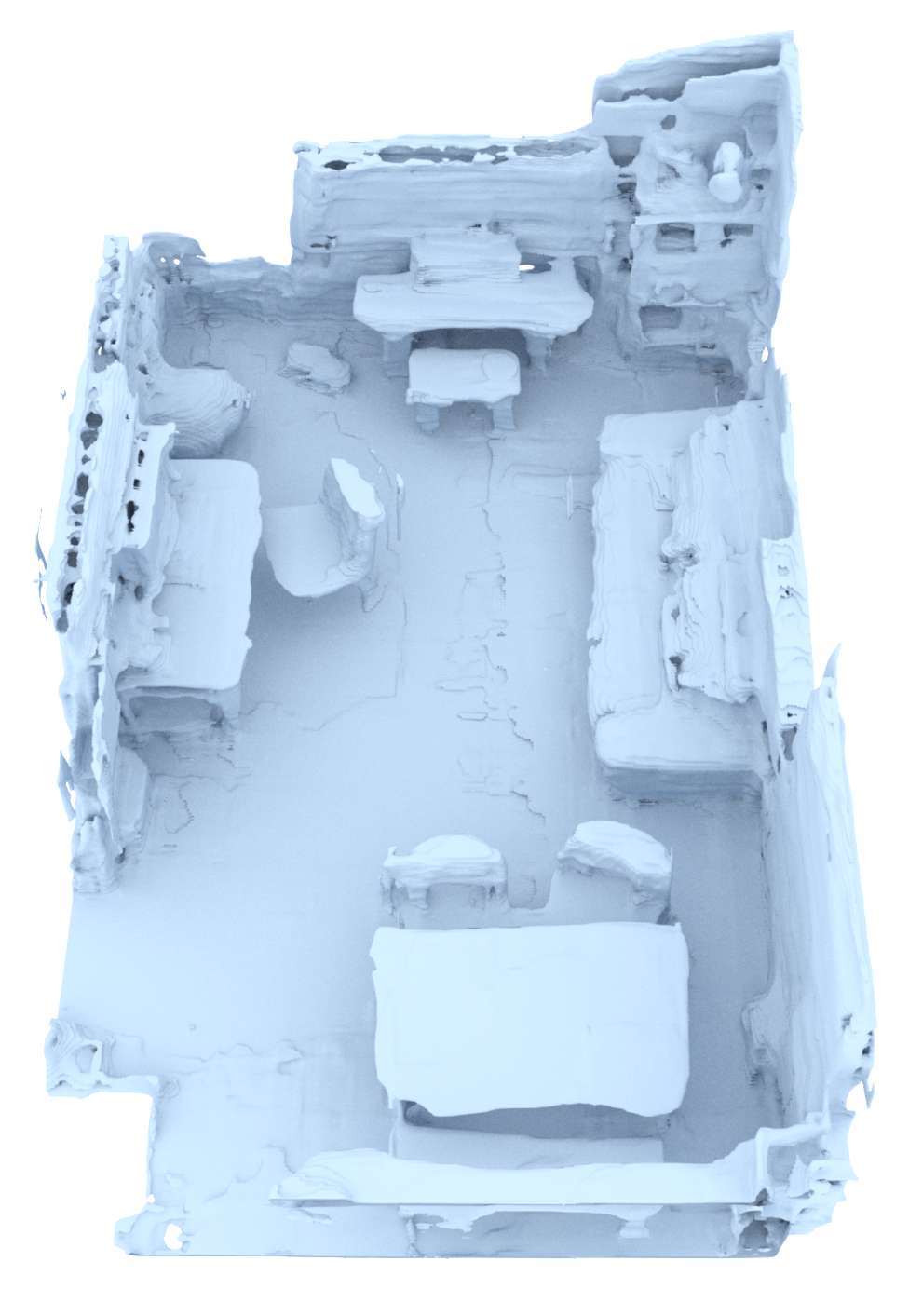}} &  \centered{\includegraphics[width=\w\textwidth]{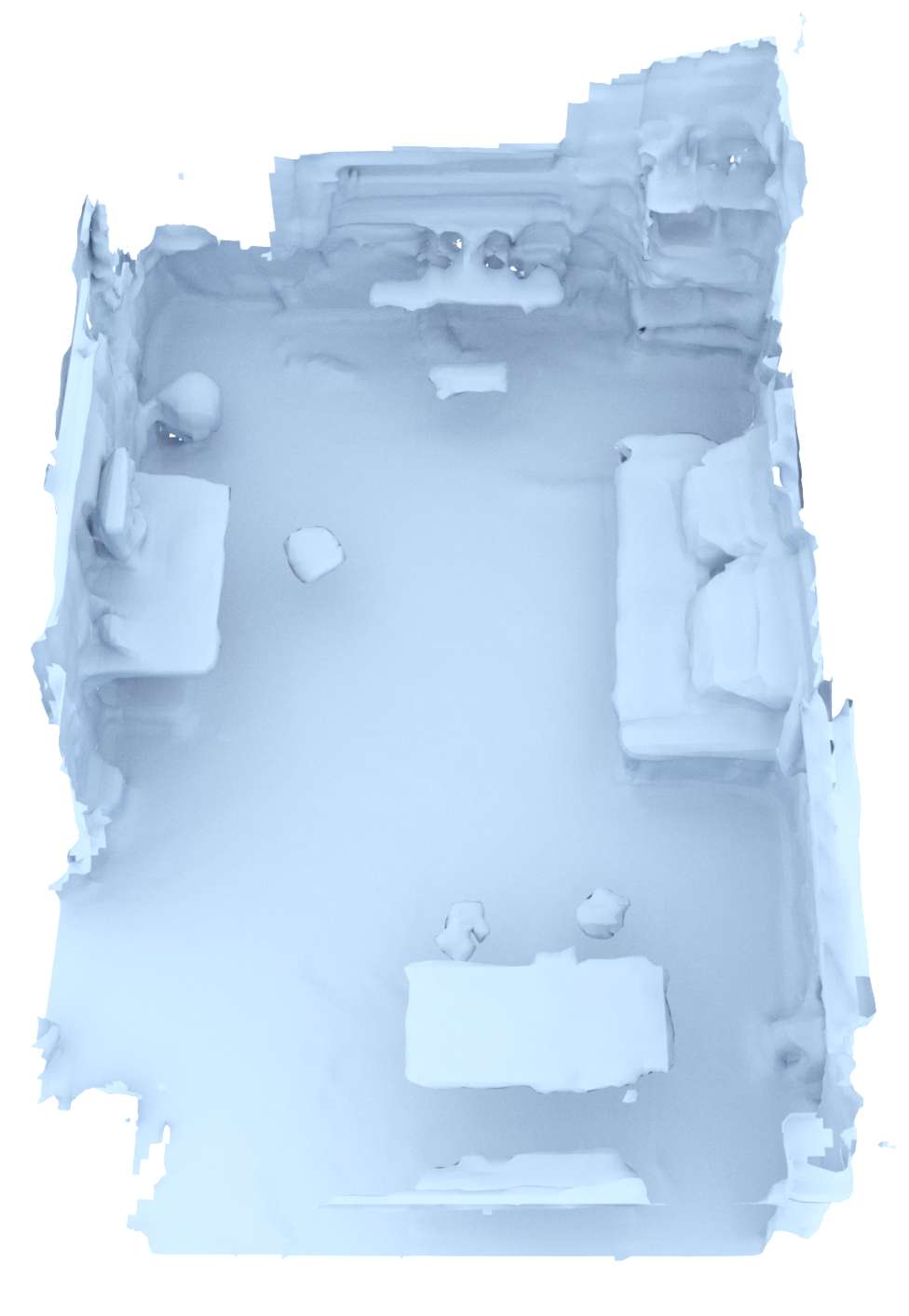}} &  \centered{\includegraphics[width=\w\textwidth]{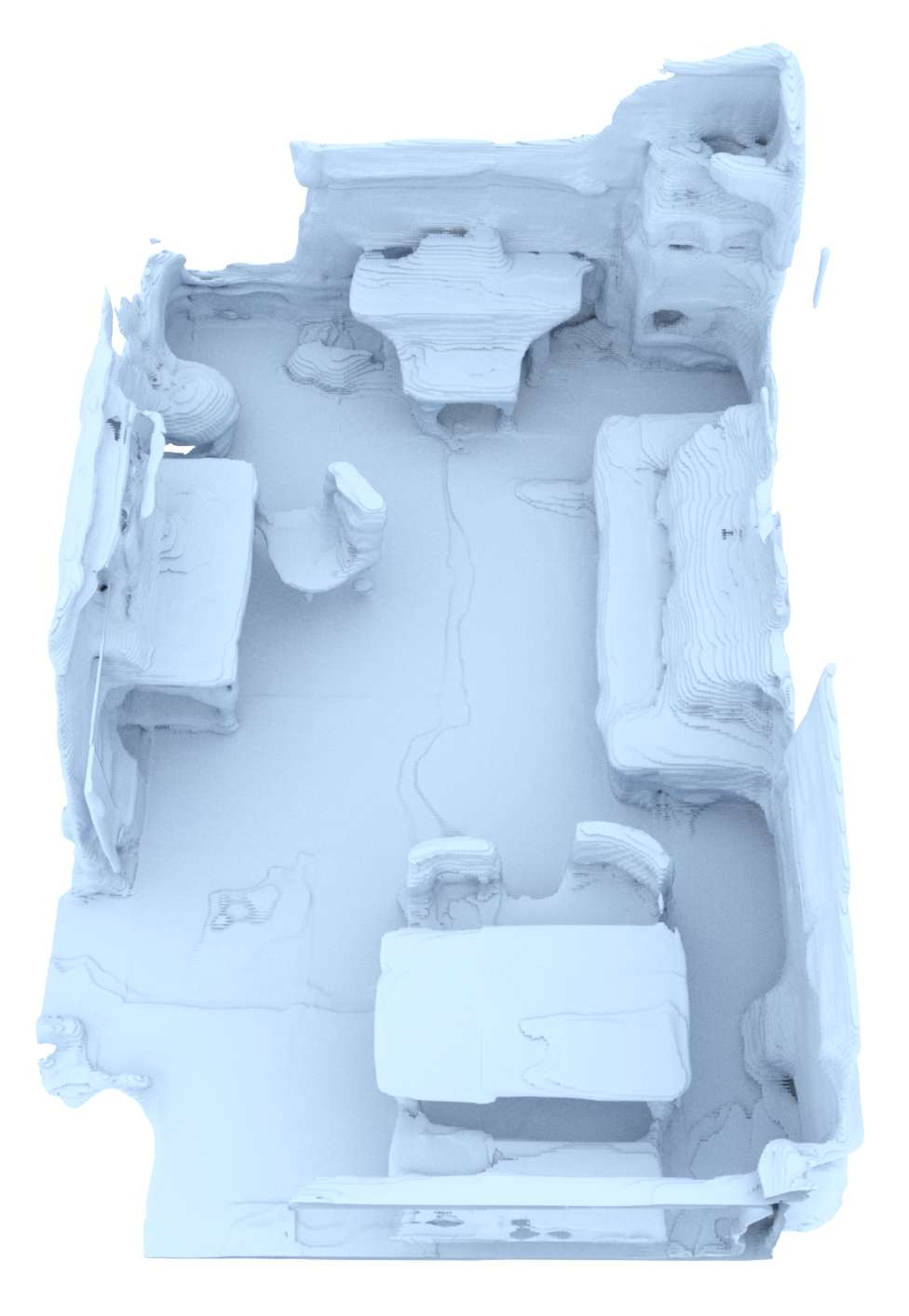}} \\
\end{tabular}
\end{center}
\caption{Qualitative results on our evaluation set. Translation noise $\sigma_T$ is applied on the $x$ and $y$ axes.}
\label{fig:qualitative-shift}
\vspace{-0.5cm}
\end{figure*}

From the quantitative and qualitative results, we can see that our models are significantly better at preserving the completeness of the reconstruction from noisy input. While accuracy and completeness are comparable for low levels of noise, our method shows significantly increased robustness in the presence of pose errors. Most importantly for use in robotic planning, our method is always able to faithfully capture the underlying geometry, shown in the high recall. This is particularly pronounced for objects of interest, \eg when the floor is excluded. Note that we train our networks with a 1 m voxel configuration and generate the 0.5 m voxel resolution maps without retraining, thus demonstrating the flexibility of our method.

\begin{figure}[h!]
\begin{center}
\begin{tabular}{@{}c@{\hspace{2em}}c@{}}
\includegraphics[width=\w\textwidth]{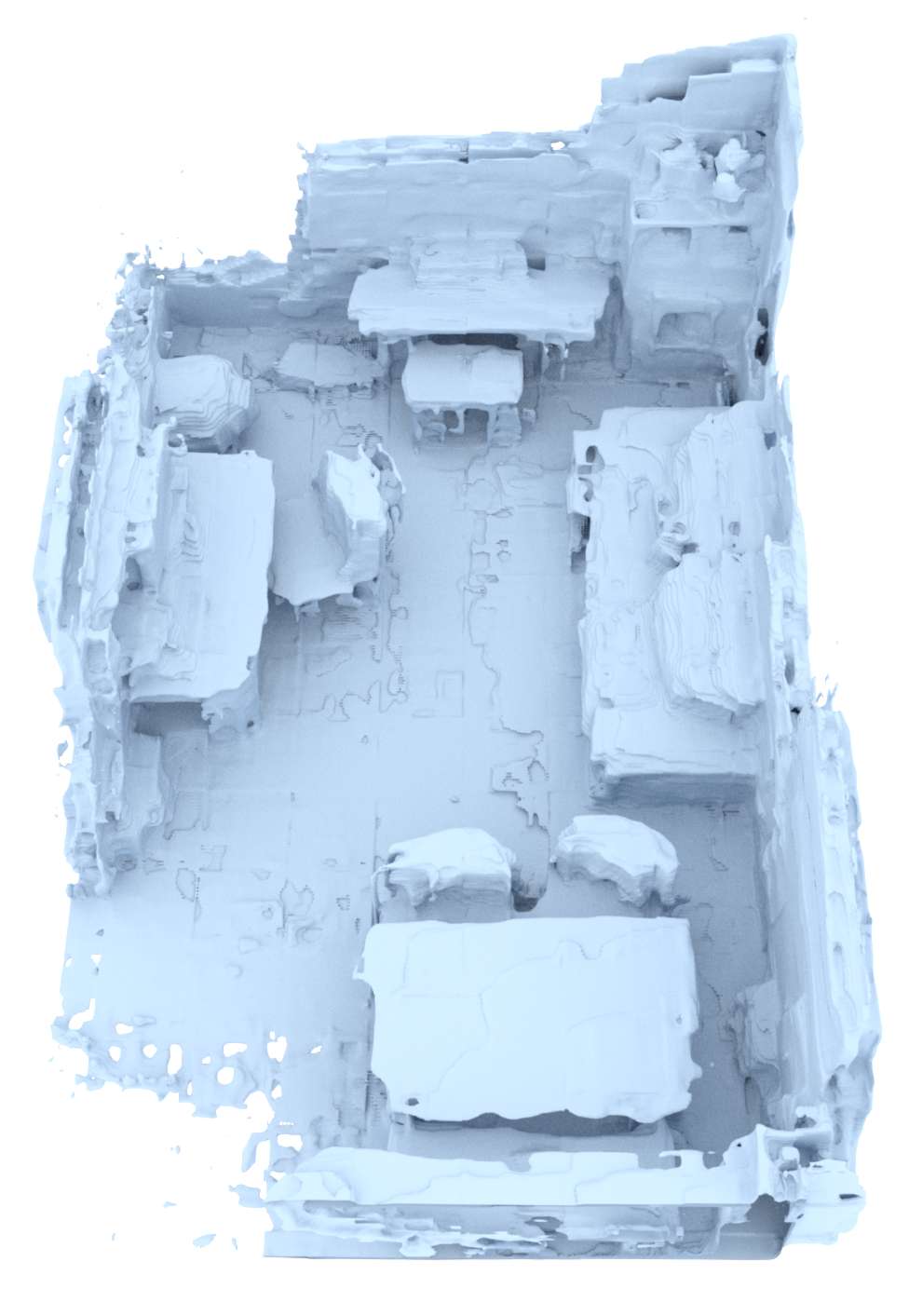} &  \includegraphics[width=\w\textwidth]{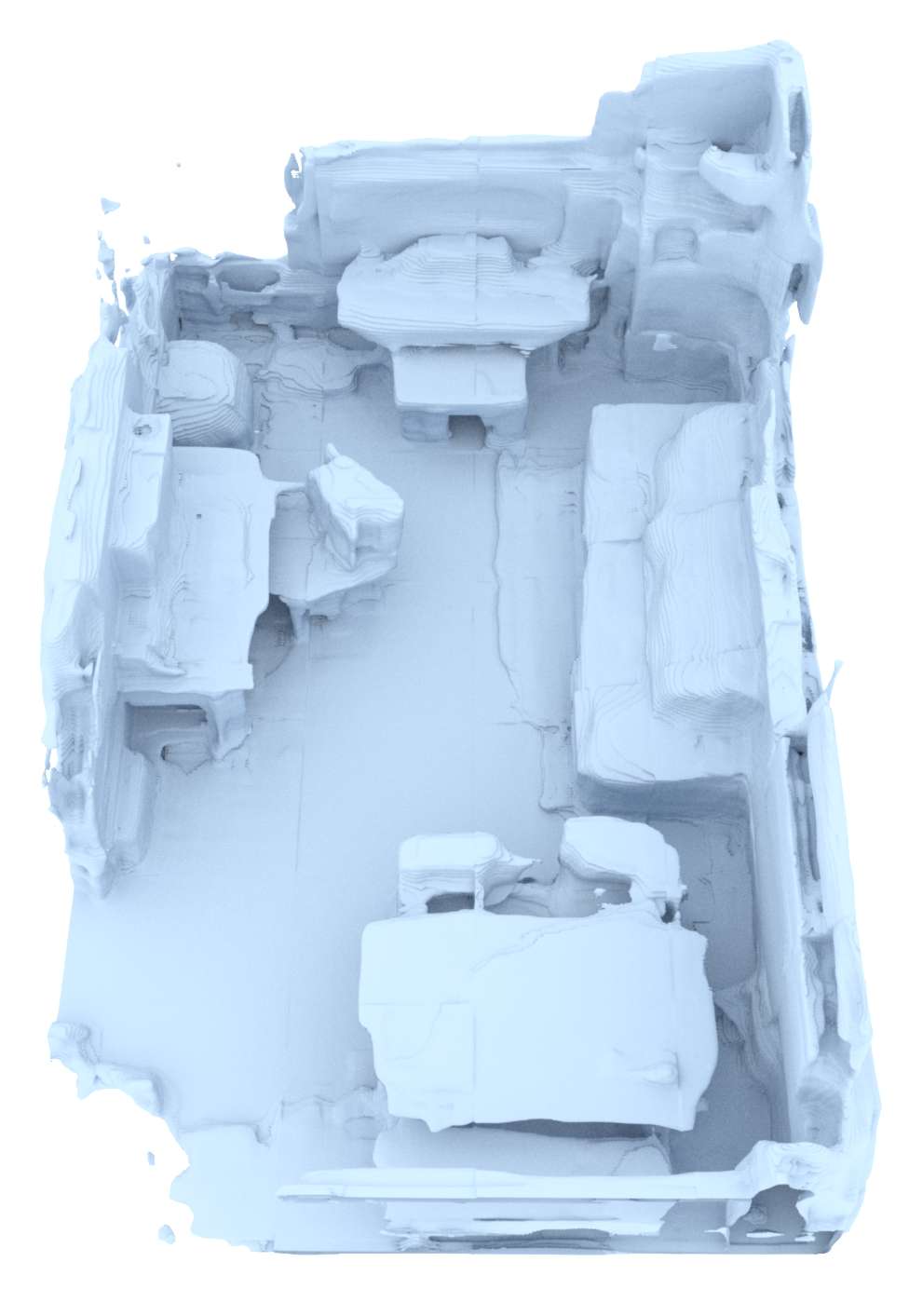} \\
\small{0.5 m voxel} & \small{1 m voxel} \\

\end{tabular}
\end{center}
   \caption{Reconstruction results using the sliding window method (as in~\cite{peng2020convolutional}) applied to accumulated noisy inputs. Simply processing accumulated noisy inputs results in inaccurate meshes with overly thick structures.}
  \vspace{-1em}
   \label{fig:trivial}
\end{figure}

We also present the reconstructed meshes using a static fusion method, \ie encoding and decoding is performed in a sliding window manner after all input point clouds are accumulated, as in 
~\cite{peng2020convolutional}. The performance of the static approach using our encoder and decoder with $\sigma_T = \SI{0.075}{m}$ is shown in Tab.~\ref{tab:trivial}. Without our fusion strategy, all of the noise is encoded into the latent codes, resulting in inaccurate meshes with overly thick structures, visible in Fig.~\ref{fig:trivial}. In contrast, our method captures the estimated statistics of input data and results in more accurate reconstructions.

\vspace{-2mm}
\begin{table}[h!]
\begin{center}
\resizebox{0.47\textwidth}{!}{%
\begin{tabular}{|c||c|c||c|c||c|c||c|c|}
    \hline
     Voxel & \multicolumn{8}{c|}{\emph{\textbf{Left:}} Static fusion~\cite{peng2020convolutional}. \emph{\textbf{Right:}} Ours.} \\
    \cline{2-9}
    size & \multicolumn{2}{c||}{\emph{Accuracy} (m)} & \multicolumn{2}{c||}{\emph{Completeness} (m)} & \multicolumn{2}{c||}{\emph{Recall}} & \multicolumn{2}{c|}{\emph{Recall*}}\\
     \hline
      0.5 m & 0.0669 & \textbf{0.0503} & 0.0416 & \textbf{0.0249} &  0.690 & \textbf{0.876} & 0.539 & \textbf{0.809} \\ 
      1 m & 0.0789 & \textbf{0.0456} &  0.0573 & \textbf{0.0286} & 0.579 & \textbf{0.849} &0.399  & \textbf{0.768} \\ 
      \hline
    \end{tabular}}
    \end{center}
    \vspace{-2mm}
    \caption{Comparison of static fusion \cite{peng2020convolutional} using our encoder and decoder and our sequential method. $\sigma_T=0.075$ m.}
    \label{tab:trivial}
\end{table}
\vspace{-0.5em}

\boldparagraph{Pointcloud sparsity and runtime} Next, we evaluate the performance of our models given different levels of input point cloud sparsity by feeding sub-sampled point clouds per scan into our model. We also report the encoding speed of our PyTorch implementation evaluated on a machine with an NVIDIA GeForce GTX 1650 GPU (4GB) and an Intel i7-9750H@2600GHz CPU. Tab.~\ref{tab:sparsity} shows the results with different levels of input sparsity.

\begin{table}[h!]
\begin{center}
\resizebox{0.47\textwidth}{!}{%
\begin{tabular}{|c||c|c||c|c||c|c||c|c|}

\hline
 \multirow{3}{*}{Input \%} & \multicolumn{8}{c|}{\emph{\textbf{Left:}} 0.5 m voxel. \emph{\textbf{Right:}} 1.0 m voxel.} \\
\cline{2-9}
 & \multicolumn{2}{c||}{\emph{Encoding}} & \multicolumn{2}{c||}{\multirow{2}{*}{\emph{Accuracy} (m)}} & \multicolumn{2}{c||}{\multirow{2}{*}{\emph{Completeness} (m)}} & \multicolumn{2}{c|}{\multirow{2}{*}{\emph{Recall}}} \\
 & \multicolumn{2}{c||}{\emph{speed (FPS)}} & \multicolumn{2}{c||}{}& \multicolumn{2}{c||}{} & \multicolumn{2}{c|}{} \\
\hline
5 & 5.2 & 11.7 & \textbf{0.0208} & \textbf{0.0281} & \textbf{0.0129} & \textbf{0.0181} & 0.988 & 0.969 \\
10 & 4.7 & 11.4 & 0.0225 & 0.0292 & 0.0135 & 0.0187 & 0.990 & \textbf{0.970} \\
50 & 2.9 & 6.3 & 0.0255 & 0.0326 & 0.0151 & 0.0206 & 0.992 & 0.965 \\
100 & 1.8 & 3.1 & 0.0263 & 0.0335 & 0.0157 & 0.0213 & \textbf{0.993} & 0.961 \\

\hline
    \end{tabular}}\\
    \end{center}
    \caption{Performance with varying levels of input point cloud sparsity. Our system achieves similar reconstruction performance even for sparse inputs, enabling it to run in real-time and with low resolution sensors. The reconstruction quality appears slightly lower for high input percentages as we tune the threshold $\tau_{occ}$ for 10\% of inputs. This can be improved by adjusting $\tau_{occ}$.}
    \label{tab:sparsity}
\end{table}

Despite being trained on 10\% input data, due to the local pooling operations that aggregate the features from $N$ point clouds into a 3D grid with dimension $24 \times 24 \times 24$, our method shows strong robustness w.r.t sparse input data and obtains similar reconstruction results given sparser inputs. It further generalizes well to more input points with only a slight drop in performance. Since the density of occupancy prediction is not directly affected by input sparsity, our models suffer less from reduced surface density and smoothness that can appear in TSDFs given sparse input. This enables our method to achieve good reconstruction results in real-time and makes it amenable to cost-efficient low resolution sensors. Reconstruction comparisons with extremely sparse input are provided in the supplementary. 

We report the speed of decoding latent codes in series, as shown in Tab.~\ref{tab:decoding-speed}. In our implementation, we directly output free space when a voxel has not received any input point cloud. Thus, the decoding voxel~/~s can be faster than the reported speed in Tab.~\ref{tab:decoding-speed} depending on the layout of the scene. It can be accelerated further by processing voxels in a batch. The encoding and decoding speeds can give an overall operational frame rate of $2.4-2.6$ FPS when querying $25^3$ and $50^3$ points/voxel for \SI{0.5}{m} and \SI{1.0}{m} voxels, respectively\footnote{Decoding is performed every frame, thus the minimum overall FPS.}.

\begin{table}[h!]
\centering
\begin{center}
\setcellgapes{1pt}\makegapedcells
\resizebox{0.47\textwidth}{!}{%
\begin{tabular}{|c|c|rrrrr|}
\cline{3-7}
\multicolumn{2}{c|}{} & \multicolumn{5}{c|}{\emph{\textbf{Query points~/~voxel}}} \\ 
\multicolumn{2}{l|}{}  & $\mathit{25^3}$ & $\mathit{50^3}$ & $\mathit{60^3}$ & $\mathit{100^3}$ & $\mathit{110^3}$ \\
\hline
\multirow{3}{1.5cm}{\centering\textbf{Decoding speed}} & \emph{Voxel~/~s} & 136.0 & 35.7 & 27.0 & 4.2 & 3.4   \\\cline{2-7}
 & \emph{FPS (0.5 m voxel)}\footnotemark  & 5.0 & 1.3 & 1.0 & - & -    \\ \cline{2-7}
 & \emph{FPS (1.0 m voxel)}\footnotemark[\value{footnote}] & 12.6 & 3.3 & 2.5 & 0.4 & 0.3  \\
\hline
    \end{tabular}}\\
    \end{center}
    \caption{Decoding speed given different query point densities.}
    \label{tab:decoding-speed}
\end{table}

\vspace{-1em}

\begin{figure*}[ht!]
\begin{center}
\begin{tabular}{@{}c@{\hspace{0.2em}}c@{\hspace{0.2em}}c@{}}
\includegraphics[width=0.33\textwidth]{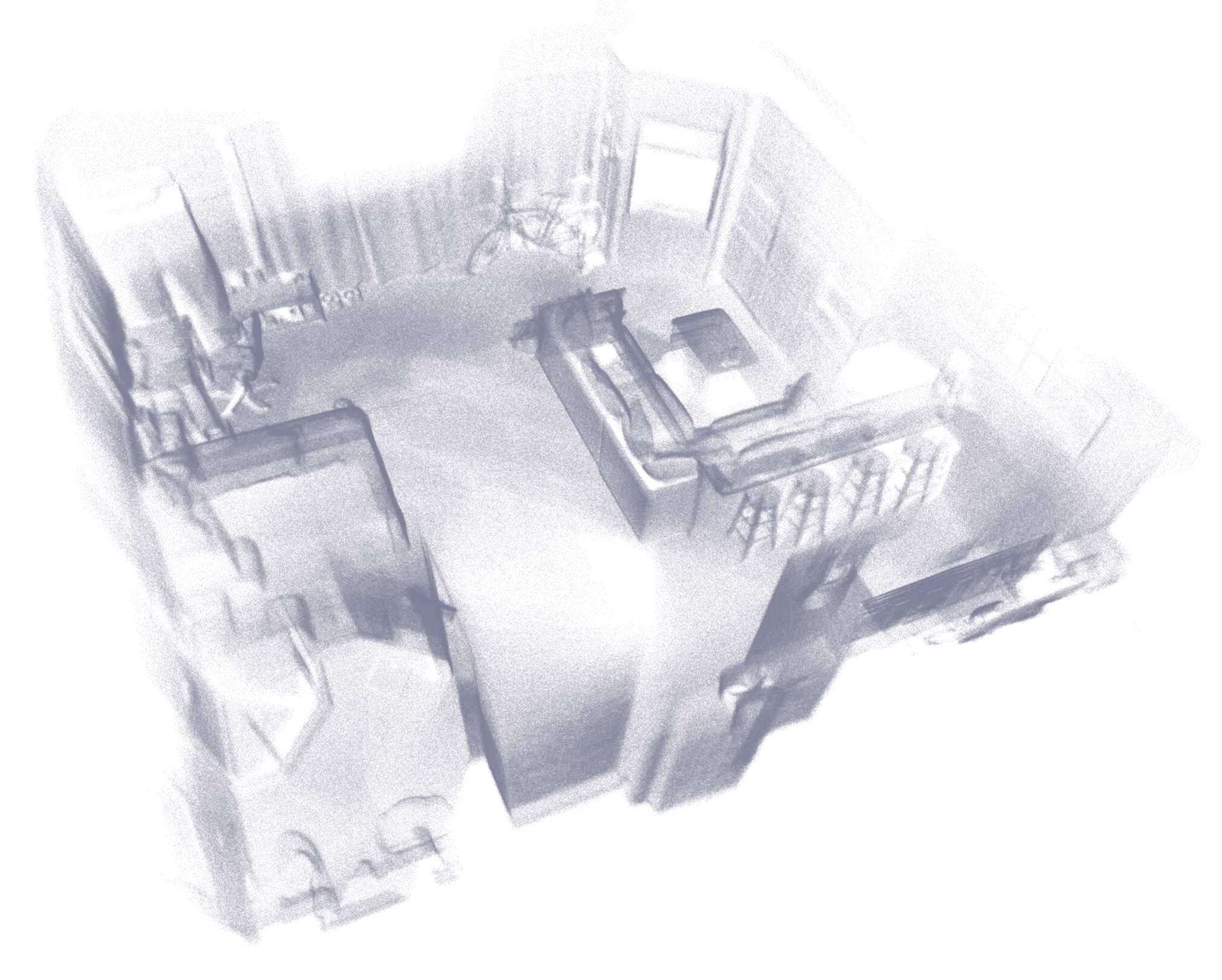} & \includegraphics[width=\ww\textwidth]{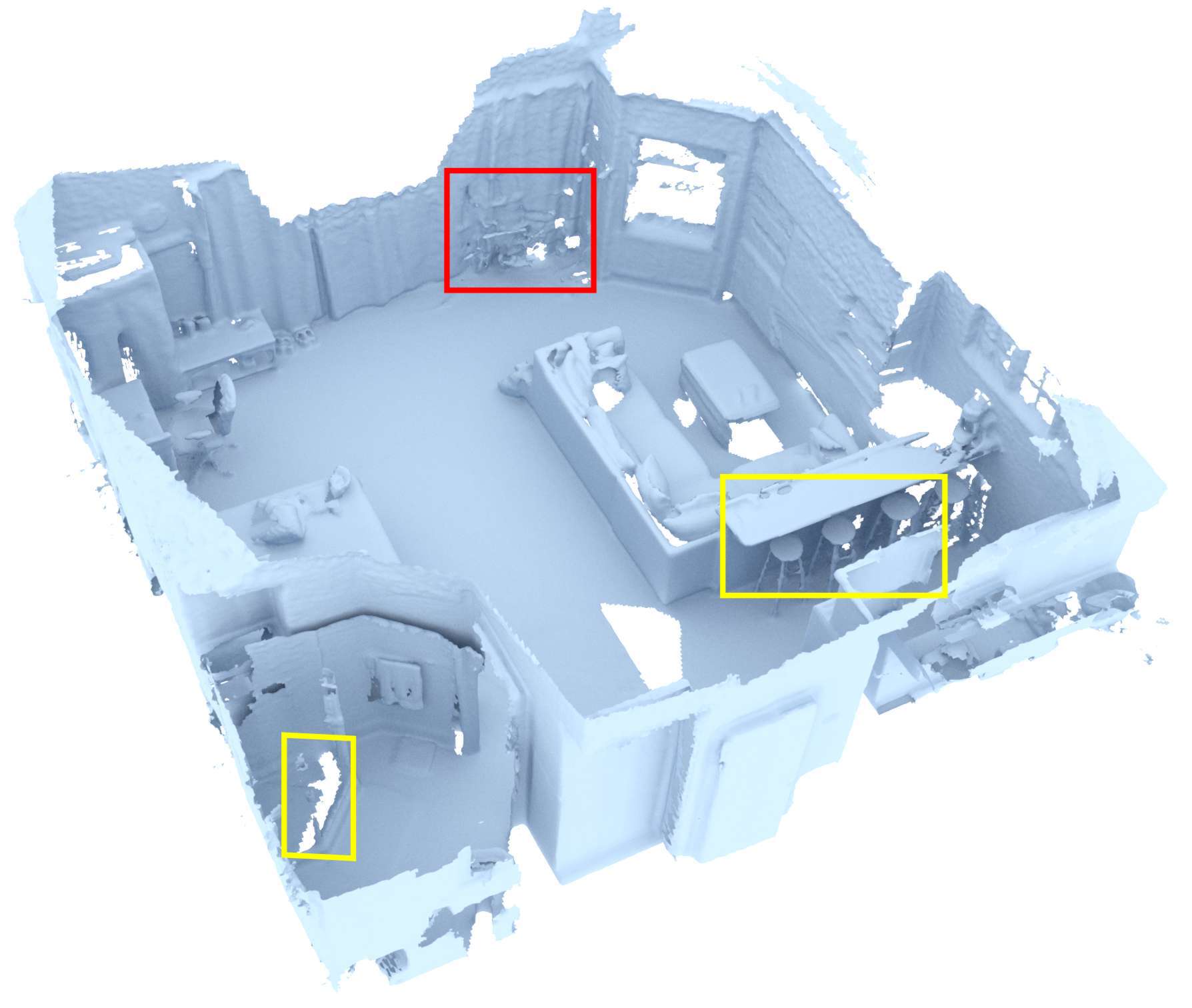} & \includegraphics[width=\ww\textwidth]{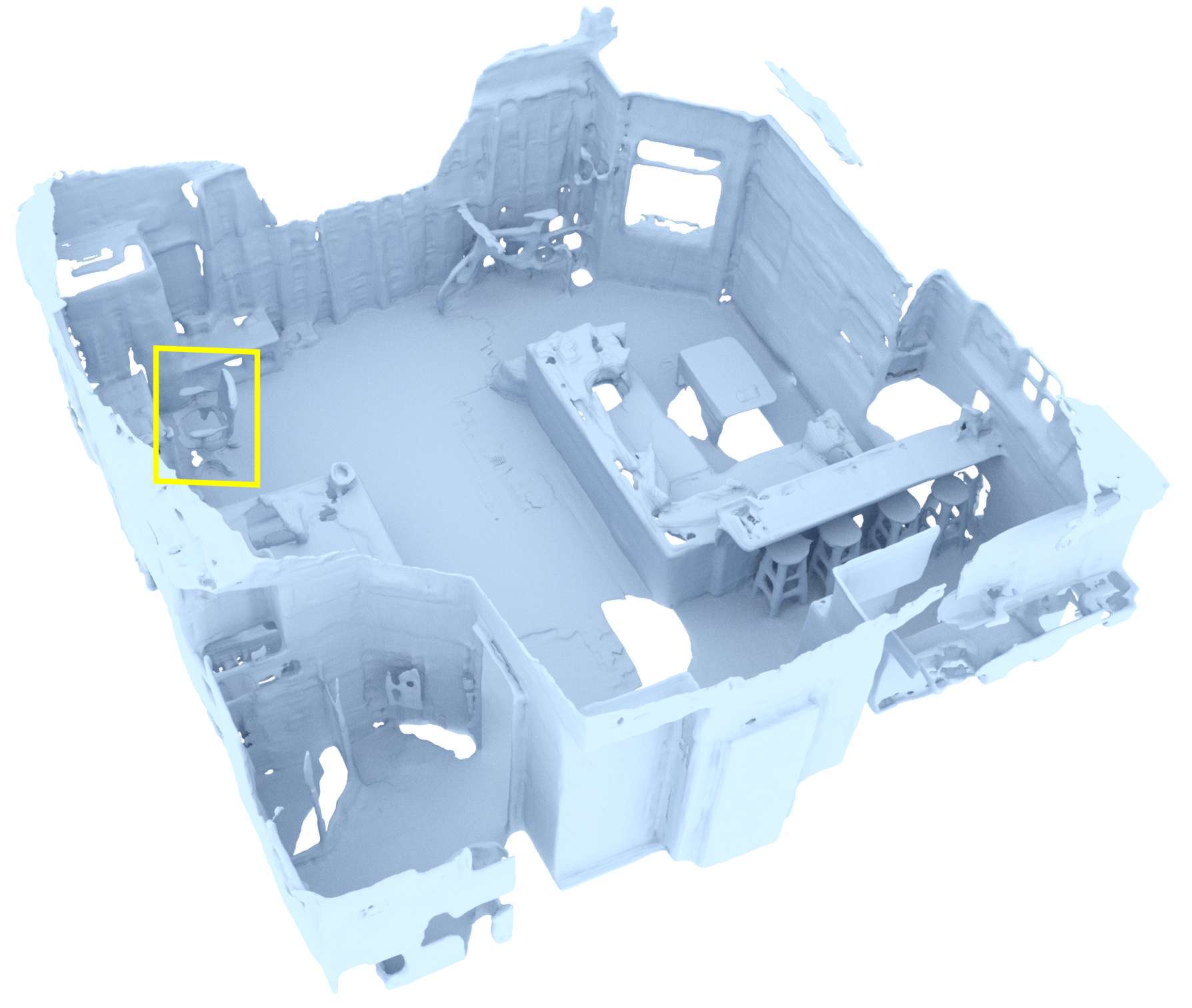} \\
\small{Sampled accumulated inputs} & \small{TSDF (0.02 m)} & \small{Ours (0.5 m)} \\

\end{tabular}
\end{center}
\caption{Qualitative results on ScanNet dataset. \emph{Red:} Failed reconstruction. \emph{Yellow:} Incomplete reconstruction.}
\label{fig:qualitative-scannet}
\vspace{-0.3cm}
\end{figure*}

\boldparagraph{CPU-only configuration} To allow our system to be employed on a compute-constrained mobile device, such as a Micro Aerial Vehicle (MAV), we demonstrate that it can run in real-time using only a CPU. We use the 1 m configuration, query dimension $d_q = d_V$, $25^3$ query points per voxel, and $5\%$ of input sparsity. Additionally, we do not update a latent code if its corresponding input volume receives less than 5\% of the input points. 
The reconstruction metrics and inference speed evaluated on an Intel i7-9750H@2600GHz CPU shown in Tab.~\ref{tab:cpu} highlight that our method can be readily employed in mobile robotics.

\begin{table}[h!]
\begin{center}
\setcellgapes{1pt}\makegapedcells
\resizebox{0.47\textwidth}{!}{%
\begin{tabular}{|c|c|c|c|c|c|}
\hline
\textbf{Encoding} & \multicolumn{2}{|c|}{\textbf{Decoding}} & \textbf{Accuracy} & \textbf{Completeness} & \textbf{Recall} \\ \hline
4.7 FPS & \unitfrac[15.6]{voxel}{s} & 2.1 FPS\footnotemark[\value{footnote}] & 0.0257 m & 0.0192 m & 0.953 \\
\hline
    \end{tabular}}\\
    \end{center}
    \caption{CPU-only configuration runtime and evaluation metrics.}
    \label{tab:cpu}
\end{table}

\footnotetext{Average speed when decoding all updated voxels every frame.}

\subsection{ScanNet}

To investigate the ability of our method to generalize to out-of-domain data, we evaluate it on ScanNet~\cite{dai2017scannet}, an RGBD dataset of real-world indoor scenes captured by a Structure sensor mounted on an iPad. 
Since no accurate ground-truth reconstruction is provided, only qualitative comparisons are presented in Fig.~\ref{fig:qualitative-scannet}. Similar to the in-domain evaluations, we observe high reconstruction quality and better object perseverance than TSDF, showing the capability of our model to generalize to scenes with different layouts and sensor characteristics. More results are provided in the supplementary materials.

\section{Discussion}
\label{sec:discussion}

\boldparagraph{Reconstruction} From the experiments in Sec.~\ref{sec:experiments}, we can see a contrasting behavior between our method and TSDF. Given inputs subject to high state estimation errors, TSDF tends to produce a reconstruction thinner than the ground truth and fails to reconstruct objects with thin geometries. In contrast, our method captures the overall statistics of the input in a large spatial context and can better preserve object existence. 
High recall shows that our method is able to faithfully capture the underlying structure, albeit with occasional thickening. However, this is preferable to vanishing surfaces for safe navigation.

\boldparagraph{Fusion strategy} Through our experiments, we show that our fusion strategy (Eq.~\ref{eq:pred}) can combine multiple latent codes of incomplete observations into a latent code of a more complete observation. Our conjecture to this ability is that the sequence of 3D convolutions in our encoder converts the input into a specific high-dimensional coordinate in latent space, as frequently is seen in autoencoder networks. 
The transformation $h_{\theta_{f}}$ in Eq.~\ref{eq:pred} then corrects the coordinate given from simply averaging latent codes into the desired coordinate of fused latent code. This ensures that our fused data can be both efficiently compressed via averaging and remain in the support domain of the pre-trained shape representation networks. 

\boldparagraph{Limitations} From Eq.~\ref{eq:pred}, it is evident that the represented shape of a voxel can be dominated by frequent repeated measurements.
To tackle this, the structure of our approach is amenable to many approaches such as key-framing or moving average integration, similar to \cite{oleynikova2017voxblox}, which gives more weight to recent observations. However, we leave the exploration of such methods for future work.

\boldparagraph{Reproducibility} Upon re-training our encoder, decoder, and fusion network from scratch, we observe that the threshold $\tau_{occ}$ that defines whether a query point is free has to be re-adjusted. Also, we find that a different total loss on convergence is observed during fusion training. This can result from different latent space landscapes being reached each time the encoder and decoder are trained from scratch.

\section{Conclusion}
\label{sec:conclusion}

We introduced a novel method for online volumetric mapping based on neural implicit representations that scales to arbitrary size environments and can incrementally update its geometric information. We formulate a method to integrate multiple neural implicit representations from sequential partial observations into a more complete representation. By fusing data in a formation of coarse voxels directly in latent space, our method leverages shape priors and captures local geometric context over temporally independent frames. We show in thorough experimental validation that this approach enables robust mapping in the presence of state estimation errors. Our system generalizes to a large variety of configurations and low-resolution range sensors and can operate in real-time even in CPU-only configurations. We make our system available as open source.

\vspace{0.5em}

\noindent{\small\textbf{Acknowledgements} \quad This work was supported by funding from the Microsoft Swiss Joint Research Center and the European Union’s Horizon 2020 research and innovation programme under grant agreement No 101017008.}

{\small
\bibliographystyle{ieee_fullname}
\bibliography{egbib}
}

\end{document}